\newcommand{\home}{\empty}
\newcommand{\wt}{\widetilde}
\documentclass[12pt]{article}
\usepackage{\home razb}

\newlength{\longeqmarginwidth}
\newlength{\longeqwidth}
\newlength{\longeqskiplength}
\newcommand{\Vzero}[2]{\mathord{\stackrel{\textstyle\kern1pt\circ}
{\smash V\vbox to6pt{}}}\vphantom{V}_{#1}^{#2}}

\newcommand{\of}[1]{\left( #1 \right)}
\newcommand{\absvalue}[1]{\left| #1 \right|}
\newcommand{\set}[2]{\left\{\hspace{0.2ex} #1 \left|\: #2
\right. \right\}}

\newcommand{\df}{\stackrel{\rm def}{=}}

\newcommand{\prob}[1]{ {\bf P}\! \left[ #1 \right] }
\newcommand{\expect}[1]{ {\bf E}\! \left[ #1 \right] }

\newcommand{\indexpect}[2]{ {\bf E}_{#1}\! \left[ #2 \right] }
\newcommand{\indprob}[2]{ {\bf P}_{#1}\! \left[ #2 \right] }

\newcounter{operator}

\usepackage{epsfig}
\usepackage{amsfonts}
\usepackage{amsmath}
\title{Improved Convergence Guarantees for Shallow Neural Networks}
\author{Alexander Razborov\thanks{University of Chicago, USA,
{\tt razborov@uchicago.edu}, Toyota Technological Institute at Chicago
and Steklov Mathematical Institute, Moscow, Russia.}}
\begin{document}
\maketitle

\begin{abstract}
We continue a long line of research aimed at proving convergence of depth 2
neural networks, trained via gradient descent, to a global minimum. Like in
many previous works, our model has the following features: regression with
quadratic loss function, fully connected feedforward architecture, RelU
activations, Gaussian data instances and network initialization,
adversarial labels. It is more general in the sense that we allow both
layers to be trained simultaneously and at {\em different} rates.

Our results improve on state-of-the-art \cite{OyS} (training the first
layer only) and \cite[Section 3.2]{Ngu} (training both layers with Le Cun's
initialization). We also report several simple experiments with synthetic
data. They strongly suggest that, at least in our model, the convergence
phenomenon extends well beyond the ``NTK regime''.
\end{abstract}

\section{Introduction} \label{sec:intro}

Deep neural networks are remarkably successful at defying common wisdom
inherited from the statistical learning theory and especially the theory of
PAC learning. It would perhaps not entirely be an exaggeration to say that
the wonder at their ability to generalize in a (highly) overparameterized
regime is expressed in at least every other paper on the theory of deep
learning so we confine ourselves to citing two expository papers
\cite{BHM*,ZBH*} {\em specifically} devoted to this phenomenon. On more
concrete side, recent works \cite{BFT,NBS,AGN*,LiL,ADH*,SoY} offer
explanations of the ``unbelievably good generalization'' from three pairwise
disjoint perspectives.

Since there does not seem to be any reasons to expect a learning system to
perform on the test examples {\em better} than on the training examples, an
even more basic theoretical question is why an algorithm, which in this area
almost always means gradient descent or its variants, achieves zero empirical
risk. It would be fair to say that this question has met with a slightly
better degree of success, particularly in the situation described by various
authors as the ``lazy training regime'' (see e.g. \cite{COB}) or the ``neural
tangent (kernel) regime'' (see e.g. \cite{MoZ}). Notwithstanding the name,
this is the behavior described by the following (very much related)
properties:
\begin{itemize}
\item only ``a few'' instance-neuron pairs change their activation during
    training;

\item network parameters do not change ``much'' (say, in the Frobenius
    norm) during training.
\end{itemize}
These properties in particular imply that the gradient of the loss function
and hence the NTK\footnote{{\em Neural Tangent Kernel}; we remind the precise
definition below but essentially it is $J_fJ_f^\top$, where $J_f$ is the
Jacobian of the map $f$ recording the network's output on $m$ data
instances.} matrix $H_t$ also do not change much from their value at
initialization. This implies that the training trajectory is almost linear
and is driven, in the error space, by the value $H_0$ of the NTK matrix at
the initialization. In the following discussion, we will adopt the
terminology ``NTK regime'' to (loosely) describe this kind of behavior.

As a by-side remark, this regime has been widely criticized for being
irrelevant to practice, and (as far as we understand) one of the main reasons
is that the above property essentially implies that the network does not
learn any useful features during training. While this criticism does not look
to us entirely unfounded, we would like to remark that, at least for
regression problems, the evidence of convergence outside the NTK regime is
either fragmentary or empirical at best (we will make a modest contribution
to the latter in Section \ref{sec:experiments}). From the mathematical point
of view, the best, and often the only, way of approaching a
difficult problem is to patiently build up necessary tools by understanding its
toy cases which in our context amounts to the NTK regime.

\bigskip

Previous work on the subject displays quite a diverse array of various
assumptions on the activation and loss functions, data, initialization, as
well as (also quite diverse) set of parameters featuring in the final results.
They are not easy to compare to each other. In order to get a more
coherent picture, we find it convenient to start with a "theoretical
benchmark". It seems the following captures a great deal of previous work (in
the sense that their message is not distorted much when reduced to
this scenario); we will separately mention several notable exceptions in
Section \ref{sec:other_research}.

\subsection{Set-up} \label{sec:set-up}

Feedforward fully connected neural networks of depth 2 (that is, with one
hidden layer). RelU activation function. Regression tasks with single output
and the squared loss function. Data points $(X^1,\ldots X^m)$ are sampled
uniformly and independently from the unit sphere. Weights at initialization
are Gaussian at the input level and Rademacher ($\{\pm 1\}$) at the output
level. Parameters:
\begin{itemize}
\item $n$ -- the problem dimension;

\item $S$ -- the number of neurons at the middle level ("size" of the network);

\item $m$ -- the number of training examples.
\end{itemize}

When $m\leq n$, the data is linearly separable. We will not exclude this case
from statements and proofs but in Section \ref{sec:intro} we will
assume for simplicity that $m\gg n$. We will also tacitly assume $m\leq
n^{O(1)}$; this assumption is indispensable in any work using NTK methods,
including ours.

\subsection{Previous work}

The over-parameterized region is $m\ll nS$ and, clearly, no convergence or
even representability is possible when $m\gg nS$. Several authors have
loosely conjectured that perhaps this is actually tight and the (stochastic)
gradient descent does converge in this whole (or at least close to it)
region. This conjecture is not as far-fetched as it may seem since under this
assumption alone one can make, with overwhelming probability, quite strong
conclusions at initialization like:

\begin{description}
\item[\cite{SoH}] The activation matrix $A\in \{0,1\}^{S\times m}$ at
    initialization possesses an intrinsic and purely combinatorial property
    ensuring the following. The corresponding NTK matrix is non-singular
    for {\em almost all}\, (in the Lebesque sense) choices of data
    $X\in\mathbb R^{n\times m}$.

\item[\cite{XLS,OyS,ADH2*,NMM,MoZ}] The NTK matrix w.r.t. the same data $X$
    that were used for computing the activation matrix $A$ has the minimum
    eigenvalue\footnote{We do not normalize by $S$.} $\Omega(S)$, i.e.
    well-separated from 0.
\end{description}

As for the dynamics of the gradient descent, the known results are much less
conclusive. Before briefly surveying them, let us again stress that many of
these results pertain to more general situations such as deeper networks,
less restrictive assumptions on data, more general loss or activation
functions etc. In our simplified treatment we have deliberately reduced them
to the ``common denominator'' outlined in Section \ref{sec:set-up}.

Let us first consider the case when only the first layer is trained.
\cite{DZR*} proved that convergence takes place whenever\footnote{We use the
notation $\wt O, \wt\Omega, \wt\Theta, \wt o,\wt\omega$ to hyde factors that
are poly-logarithmic in $n,S$, and their tildeless versions to hyde constant
factors. See Remark \ref{rem:asymptotic_notation} below for clarifying
remarks and examples.} $m\leq \wt o\of{S^{1/6}}$ (plug $m\mapsto S;\ n\mapsto
m;\ \lambda_0,\delta\mapsto O(1)$ in their Theorem 4.1); apparently, this can
be approved to $m\leq \wt o\of{n^{1/6}S^{1/6}}$ by examining their proof more
carefully. The paper \cite{WDW} improved the bound on the learning rate from
\cite{DZR*} using more sophisticated (adaptive) gradient methods. The
improvement $m\leq \wt o\of{n^{1/2}S^{1/4}}$ can be relatively easily
extracted from \cite{SoY} (plug $m\mapsto S;\ n\mapsto m;\ \lambda\mapsto
O(1);\ \alpha\mapsto \wt O(m/n);\ \theta\mapsto \wt O(m^{1/2}/n^{1/2})$ in
their Theorem 1.6). Finally, the paper \cite{OyS} achieved the
state-of-the-art result: convergence takes place whenever
\begin{equation} \label{eq:oymak}
m\leq \wt o\of{n^{3/4}S^{1/4}}.
\end{equation}

\smallskip
Our description may look somewhat uneventful but we would like to stress that
every new achievement along these lines required introducing new ingenious and
technically advanced ideas and methods.

Less work seems to have been done on the case when both layers are trained
simultaneously, and now the answer may depend on the normalization of weights
at the initialization even if we disregard the learning rate. The original paper \cite{DZR*} gives the same
estimate $m\leq \wt o\of{S^{1/6}}$ under the ``standard'' normalization, when
the initial weights are supposed to be of the same order (say, $\wt O(1)$) at
both layers. The paper \cite{Ngu} is mostly devoted to deep networks but in
Section 3.2 it also translates the main result to the shallow case under the
so-called LeCun's initialization. After performing slightly more careful
calculations, in our notational system this amount to the condition
\begin{equation} \label{eq:nguyen}
m\leq \min\of{\wt o\of{\frac Sn},\ \wt o\of{S^{1/2}}}.
\end{equation}

\subsection{Our contributions} \label{sec:contributions}
We work in the natural multi-rate setting in which the first layer is trained
at a rate $\eta_w\geq 0$ and the second layer is trained at possibly another
rate $\eta_z\geq 0$. This view appears to be very instructive and allows us
to easily translate results from one normalization of initial weights to
another. In particular, we choose to work in the standard normalization, when
all weights are of order $\wt O(1)$.

As a by-side remark, the recent paper \cite{VlL} expressed surprise (that we
completely share) that this simple idea seems to have been largely overlooked
in the theory of {\em deep} learning and sought to change that by providing,
among other things, significant experimental evidence. One difference between
our paper and \cite{VlL} is that the latter restricts the {\em set of steps}
in which a ``slow'' set of parameters is changed while we treat all steps
uniformly. But this difference appears to be cosmetic.

Skipping a few inessential details (see Theorem \ref{thm:main} for the
official statement), our main contribution is as follows.

\begin{theorem}[informal] \label{thm:informal}
  In the set-up described above, assume that $n,S\geq\wt\omega(1)$,
  $m\leq n^{O(1)}$ and that $m\leq\wt o(S)$. Assume
  additionally that one of the following happens:
\begin{equation} \label{eq:w_regime}
m\leq{\wt o\of{nS^{1/4}}}, \ \ \ \frac{\eta_z}{\eta_w} \leq \wt
o\of{\frac Sm},
\end{equation}

\begin{equation} \label{eq:wz_regime}
m\leq{\wt o\of{nS^{1/4}}}, \ \ \ \frac{\eta_z}{\eta_w} \geq \wt
\omega\of{\frac{m^2}{nS}}
\end{equation}
or

\begin{equation} \label{eq:z_regime}
\frac{\eta_z}{\eta_w} \geq \wt
\omega\of{1+\frac mn}.
\end{equation}
Assume also that $\eta_w,\eta_z$ are sufficiently small. Then with
probability $\geq 1-(nS)^{-\omega(1)}$ the gradient descent converges to a
global minimum.
\end{theorem}

\begin{remark} \label{rem:asymptotic_notation}
This paper uses the asymptotic notation $\wt O, \wt\Omega, \wt
o,\wt\omega$ (and their tildeless versions) perhaps more systematically and
heavily than most papers on the subject. Thus it would be appropriate
to pause for a moment and reflect on what exactly it means.

We argue about infinite sequences $\left\{\of{n^{(k)}, m^{(k)}, S^{(k)},
\eta_w^{(k)}, \eta_z^{(k)}}\right\}_{k\geq 1}$ of parameter values. Then (say)
$m\leq\wt o\of{nS^{1/4}}$ means that for any fixed $C>0$,
\begin{equation} \label{eq:asymptotic_simple}
\lim_{k\to\infty} \frac{m^{(k)}\log (n^{(k)} S^{(k)})^C}{n^{(k)}(S^{(k)})^{1/4}} = 0
\end{equation}
while the error bound $(nS)^{-\omega(1)}$ simply states that it is
asymptotically smaller than any fixed polynomial in\footnote{We do not include
$m$ as we are assuming $m\leq n^{O(1)}$ throughout the paper.} $n,S$. The
condition $n,S\geq \wt\omega(1)$ says that $S$ and $n$ are (very mildly)
balanced: $S\geq (\log n)^{\omega(1)}$ and $n\geq (\log S)^{\omega(1)}$; it
appeared in most of the previous work.

For a bit of further discussion see also Remark \ref{rem:dichotomy} below.
\end{remark}

\begin{remark}
The three cases \eqref{eq:w_regime},
\eqref{eq:z_regime}, and \eqref{eq:wz_regime} are in fact three separate statements in which we set our
hopes to converge fast on the first,  second or both layers, respectively. We
have combined them into one ``mega-theorem'' since the proofs share many
steps.
\end{remark}

\medskip
Let us now compare our result with the previous work cited above. The case
$\eta_z=0$ (that is, training the first layer only) automatically implies the
second condition in \eqref{eq:w_regime}, and the only remaining assumptions
are
\begin{equation} \label{eq:main_w}
m\leq \min\of{\wt o(S),\ \wt o\of{nS^{1/4}}}
\end{equation}
which is an improvement on \eqref{eq:oymak}.

The condition $\frac{\eta_z}{\eta_w} \leq \wt o\of{\frac Sm}$ in
\eqref{eq:w_regime} is automatically satisfied when $\eta_z=\eta_w$ (as
$m\leq \wt o(S)$), i.e. for the ``ordinary'' gradient descent. Thus our
result improves on \cite{DZR*}. As an easy calculation shows, re-normalizing
from LeCun's initialization to the standard one gives the ratio
$\frac{\eta_z}{\eta_w} = \frac Sm$. Thus the condition \eqref{eq:z_regime}
holds whenever $m\leq \min\of{\wt o(S),\ \wt o\of{n^{1/2}S^{1/2}}} $. This
gives an improvement on \eqref{eq:nguyen}.

Finally, note that when $m\leq \wt o\of{n^{1/3}S^{2/3}}$, the ranges for the
relative learning rate $\frac{\eta_z}{\eta_w}$ in \eqref{eq:w_regime} and
\eqref{eq:wz_regime} overlap. Thus, under the assumptions
$$
m\leq \min\of{\wt o(S),\ \wt o\of{n^{1/3}S^{2/3}},\ \wt o\of{nS^{1/4}}},
$$
convergence of the gradient descent is guaranteed {\em regardless} of the
relative learning rate; we only have to make sure the rates are individually small
enough. In particular, this holds regardless of the choice of normalization at
the initialization.

\medskip
Our improvements are admittedly not dramatic but our main motivation was to
develop new techniques that might turn out useful for pushing the bound
further or in other similar situations. This is what we consider our main
technical contributions.

\begin{itemize}
\item Most previous work used, explicitly or implicitly, bounds on the
    spectral norm $||X||$, where $X$ is the data matrix. We will heavily
    exploit that when $X$ is random, analogous bounds also hold for all its sub-matrices $X^J$,
    where $J$ is a subset of data instances (see \eqref{eq:X_norm}). We
    will also employ a dual lower bound on $\sigma_{\min}\of{(X^J)^\top}$
    for all sufficiently large $J$ (see \eqref{eq:X_dual} and compare with
    \cite[Assumption 2.2]{HuY}).

\item We will also need to extend the lower bound
    $\lambda_{\min}(H_0)\geq\Omega(S)$ to the uniform lower bound
    $\lambda_{\min}(H_0|_\Gamma)\geq\Omega(S)$, where $\Gamma\subseteq [S]$
    is a sufficiently large but otherwise arbitrary set of neurons and $H_0|_\Gamma$
    is obtained from $H_0$ by disregarding neurons not in $\Gamma$. It turned out surprisingly
    difficult and required a version of the Lipschitz concentration
    inequality (see the second part of Appendix
    \ref{app:ntk_initialization}).

\item For the first two parts  \eqref{eq:w_regime}, \eqref{eq:wz_regime},
    we also need to do finer analysis of the changes the activation matrix
    $A$ incurs during training. Informally speaking, there are two
    different ways the network may try to escape the NTK regime: by
    accumulating a large overall number of changes or a relatively large
    number concentrated on a small set of inputs $J$. Attempts of the first
    kind will be prevented by the assumption $m\leq{\wt o\of{nS^{1/4}}}$
    while the second kind will be taken care of by $m\leq\wt o(S)$. See
    Appendix \ref{app:ntk} for details.

\item Since in case of RelU activations the gradient is discontinuous, we
    have to take care of the unpleasant situation when $A$ and hence the
    Jacobian $J_F$ of the feature map significantly change during one step.
    The previous work (see e.g. \cite[Appendix G]{OyS}) did it by comparing
    both Jacobians to the Jacobian at initialization but it does not work
    any longer with our relaxed assumptions. Instead, we adopt to our
    purposes the beautiful invariant from \cite{ACH,DHL} concerning the
    behavior of weights associated to an individual neuron. See Appendix
    \ref{app:invariant} for details.
\end{itemize}

In Section \ref{sec:experiments}, we will also report several simple
experiments with synthetic data. We defer further discussion to that section;
for now let us just remark that one strong conclusion we draw from these
experiments is that there should be very significant room for further theoretical improvements, possibly (almost) all the way up to the representability barrier
$m\sim nS$.

\subsection{Related research} \label{sec:other_research}

Our attempt at the uniformization has inevitably left behind some very
interesting and somewhat related work. Very briefly:

\smallskip
{\bf Deeper networks.} Several papers (notably \cite{ALS,ZoG}) are
specifically devoted to the NTK regime in the context of deeper networks.
When scaled down to shallow networks, however, the results do not seem to be
as good as those obtained with methods specifically tailored to that
situation.

\smallskip
{\bf Classification problems.} Quite a bit of similar and generally stronger
results have been obtained for classification problems, both binary and
multi-classification. See e.g. \cite{BGM*,ZCZ*,JiT,CCZ*,FCB,LyZ}. While some
methods and approaches are shared with the regression problems, many
techniques seem to be specific to classification.

\smallskip
{\bf Smooth activation.} This case tends to be easier than the case of RelU,
and many technical difficulties disappear. For strong results in that
direction not covered above see e.g. \cite{HuY,SRP*,BAM}.

\section{Preliminaries and the main result}

\subsection{Notation}

We let $[n]\df \{1,2,\ldots,n\}$ and let ${V\choose n}$ be the collection of
all $n$-element subsets of $V$. For a real number $t$, $\lfloor t\rfloor$ is
its integer part and $\{t\}\df t-\lfloor t\rfloor$ is its fractional part.
All vectors and matrices are real, and all vectors are column vectors unless
otherwise noted. Matrices are generally denoted by upper case Latin letters
like $A,B,H,W,X$ and vectors are typically denoted by lower case letters. We
let $\mathbb R^{n\times m}$ be the space of $n\times m$ matrices. For $M\in
\mathbb R^{n\times m}$ and $j\in [m]$, we let $M^j$ denote its $j$th column
vector, and for $i\in [n]$, $M_i$ is its $i$th row vector. More generally,
for $J\subseteq [m]$, $M^J$ is the corresponding $n\times |J|$ matrix, and
similarly for $M_I$. For $z\in\mathbb R^S$, $\text{diag}(z)\in \mathbb
R^{S\times S}$ is the corresponding diagonal matrix.

The symbol $\circ$ stands for the Halliard (entryism) product of
vectors/matrices of the same size. The symbol $\ast$ will denote the {\em
column-wise} Khatri-Rao product: for $A\in \mathbb R^{S\times m}$ and $X\in
\mathbb R^{n\times m}$, $(A\ast X)\in \mathbb R^{Sn\times m}$ and $(A\ast
X)_{(\nu i),j}\df A_{\nu j}X_{ij}$. We will denote the standard Euclidean
norm by $||x||$. The unit sphere $S^{n-1}\subseteq \mathbb R^{n}$ is given by $S^{n-1}\df\set{x\in\mathbb R^n}{||x||=1}$; more generally,
$r\cdot S^{n-1}\df\set{x\in\mathbb R^n}{||x||=r}$. The spectral norm $||M||$
of $M\in \mathbb R^{n\times m}$ is $||M||\df\max_{\xi\in S^{m-1}}||M\xi||$, its
minimum singular value is $\sigma_{\min}(M)\df\min_{\xi\in S^{m-1}}||M\xi||$ and
the Frobenius norm is $||M||_{\text{\tiny F}}\df\of{\sum_{i\in [n]}\sum_{j\in
[m]}M_{ij}}^{1/2}$. We let $||x||_\infty,\ ||M||_\infty$ denote the maximum
absolute value of an entry.

When $H$ is a PSD (positive semi-definite) matrix, we will usually write
$\lambda_{\min}(H)$ for $\sigma_{\min}(H)$; thus, for any real matrix $M$,
$\sigma_{\min}(M)=\lambda_{\min}(M^\top M)^{1/2}$. For two symmetric matrices $M,N$
of the same size, $M\succeq N$ means that $M-N$ is PSD.

We let $\expect{\ast}$ be the expectation of a real random variable,
$\text{Var}(\ast)$ be the variance of a (one-dimensional) random
variable and $\prob E$ be the probability of an event $E$. $\mathbf 1(E)$ is
the characteristic function of $E$. When we randomize over a part of the
sample space, this will be indicated by the corresponding subscript, like
$\indexpect{W_0}{\ast}$ or $\indprob{z}{\ast}$. Let $||\zeta||_{\psi_2},\
||\zeta||_{\psi_1}$ be the sub-gaussian and the sub-exponential norms of a
one-dimensional $\zeta$ defined as
$$
||\zeta||_{\psi_p} \df \inf\set{s>0}{\expect{e^{(\zeta/s)^p}}\leq 2}.
$$

\subsection{Useful facts}

\noindent{\bf Shur's theorem and inequalities.} The Halliard product $M\circ
N$ of two PSD $m\times m$ matrices is again PSD and satisfies
\begin{eqnarray*}
||M\circ N|| &\leq& \max_{j\in [m]} |M_{jj}| \cdot ||N||;\\
\lambda_{\min}(M\circ N) &\geq& \min_{j\in [m]} |M_{jj}|\cdot \lambda_{\min}(N).
\end{eqnarray*}
If $A\in\mathbb R^{S\times m}$ and $X\in\mathbb R^{n\times m}$ then
$$
(A\ast X)^\top (A\ast X) = (A^\top A)\circ (X^\top X).
$$
In particular, applying Shur's inequalities with $M\mapsto A^\top A$,
$N\mapsto X^\top X$, we get
\begin{eqnarray}
\label{eq:shur}  ||A\ast X|| &\leq& \max_{j\in [m]}||A^j|| \cdot ||X||;\\
\nonumber \sigma_{\min}(A\ast X) &\geq& \min_{j\in [m]}||A^j||\cdot \sigma_{\min}(X).
\end{eqnarray}

\medskip\noindent
{\bf Norm bounds.} For $z\in\mathbb R^S$, the norm $||(\text{diag}(z)A)\ast
X||$ can be bound in two different (dual) ways:
\begin{eqnarray}
\label{eq:bound_infinity} ||(\text{diag}(z)A)\ast X|| &\leq& ||z||_\infty \cdot
||A\ast X||;\\
\label{eq:bound_diagonal}
||(\text{diag}(z)A)\ast X|| &\leq& ||z|| \cdot \max_{\nu\in S}||X\text{diag}(A_\nu)||
\leq ||z||\cdot ||A||_\infty \cdot ||X||.
\end{eqnarray}

Another handy inequality is
\begin{equation} \label{eq:handy}
||MN||_{\text{\tiny F}}\leq ||M||_{\text{\tiny F}} \cdot ||N||.
\end{equation}

\medskip\noindent
{\bf $\epsilon$-nets \cite[Chapter 5.2.2]{Ver}.}  An {\em $\epsilon$-net} in
$S^{n-1}$ is any subset $\mathcal N\subseteq S^{n-1}$ such that $\forall v\in
S^{n-1}\exists \xi\in\mathcal N(||v-\xi||\leq\epsilon)$.

\begin{fact}[\protect{\cite[Lemma 5.2]{Ver}}] \label{fct:net1} For any $\epsilon, n$ there exists an
$\epsilon$-net $\mathcal N$ in $S^{n-1}$ with $|\mathcal N|\leq \of{1+\frac
2\epsilon}^n$.
\end{fact}

\begin{fact}[similar to \protect{\cite[Lemma 5.3]{Ver}}] \label{fct:net2} Let $M\in \mathbb R^{k\times
n}$, $\sigma >0$ and $\mathcal N$ be an $\frac{\sigma}{2||M||}$-net in
$S^{n-1}$ such that $\forall \xi\in\mathcal N(||M\xi||\geq \sigma)$. Then
$\sigma_{\min}(M)\geq \frac{\sigma}{2}$.
\end{fact}

\subsection{The model} \label{sec:model}

Let $n,m,S\geq 1$. We are given $m$ data instances $X^1,\ldots,X^m\in\mathbb
R^n$ arranged as a matrix\footnote{In many works on the subject, data points
are arranged as {\em rows} of the data matrix. We have found that, while also
imperfect, the opposite convention leads to a cleaner and more natural
notation.} $X\in \mathbb R^{n\times m}$ and a {\em label vector} $y\in\mathbb
R^m$. A (shallow) neural network is given by a pair $\theta=(W,z)$, where
$W\in\mathbb R^{S\times n}$ and $z\in\mathbb R^S$. It computes as
\begin{eqnarray*}
F(W) &\df& \sigma(WX) \in \mathbb R^{S\times m};\\
f(\theta) &\df& F(W)^\top z \in \mathbb R^m,
\end{eqnarray*}
where, as usual, the RelU activation function $\sigma(x)\df \max(x,0)$ is
applied to $WX$ entryism. We define the {\em error vector} as
$$
e(\theta) \df f(\theta) -y \in \mathbb R^m
$$
and the (quadratic) {\em loss function} as
$$
\ell(\theta) \df \frac 12 ||e(\theta)||^2.
$$
Let us call $\theta=(W,z)$ {\em regular} if $WX$ does not contain zero entries
and {\em singular} otherwise. All functions just defined are polynomial in a neighborhood
of any regular $\theta$, hence the gradient $\nabla\ell(\theta)$ is well-defined for regular $\theta$. We split it as $\nabla\ell(\theta) = (\nabla^w\ell(\theta),\
\nabla^z\ell(\theta))$ in the obvious way:
$\nabla^z\ell(\theta)\in\mathbb R^S$ and $\nabla^w\ell(\theta)$ will be
treated, depending on the context, either as a long vector in $\mathbb
R^{nS}$ or as a matrix in $\mathbb R^{S\times n}$.

Assume now that we are also given initial values $\theta_0=(W_0,z_0)$ and the
{\em learning rates} $\eta_w,\eta_z\geq 0$. We
define the {\em gradient descent} $\theta_\tau=(W_\tau, z_\tau)$ ($\tau\geq
1$ an integer) as
\begin{equation} \label{eq:gradient_descent}
\begin{split}
W_\tau &\df W_{\tau -1}-\eta_w\nabla^w\ell(\theta_{\tau-1});\\
z_\tau &\df z_{\tau-1}-\eta_z\nabla^z\ell(\theta_{\tau-1}).
\end{split}
\end{equation}
There is a small caveat here since this definition makes sense only when all
points $\theta_\tau$ are regular. Fortunately, it is easy to take care of
with a simple measure-theoretical argument that we will present in Appendix
\ref{app:caveat}.

\subsection{Data, initialization and the main result} \label{sec:data}
We pick the data $X^1,\ldots,X^m$ uniformly (under the Haar measure) and
independently from $S^{n-1}$. All entries in $W_0$ are chosen from $\mathcal
N(0,1)$, independently of each other. We prefer to allow for a slightly
better flexibility in the choice of $z_0$, primarily since the literature alternates
between Gaussian and Rademacher initializations.

Recall that the notation $\wt O,\ \wt\Omega, \wt o,\wt\omega$ hides factors
that are poly-logarithmic in $\log(nS)$; see Remark
\ref{rem:asymptotic_notation} for more details. We require that $z_0$ is
independent of $W_0$, and that the entries in $z_0$ are $S$ i.i.d. copies of
a one-dimensional distribution $\zeta$ such that:
\begin{eqnarray}
\nonumber \expect{\zeta} &\df& 0;\\
\label{eq:psi2} ||\zeta||_{\psi_2} &\leq& \wt O(1);\\
\label{eq:var} \text{Var}(\zeta) &\geq& \wt\Omega(1).
\end{eqnarray}

We are now ready to state our main result.

\begin{theorem} \label{thm:main}
Let $m\geq 1$ and $n,S\geq\wt\omega(1)$ be parameters such that
  $m\leq n^{O(1)}$ and $m\leq\wt o(S)$. Let $\eta_w,\eta_z\geq 0,\ \eta_w+\eta_z>0$
be such that
\begin{equation} \label{eq:absolute_bounds}
\eta_w\leq \min\of{\wt o\of{\frac 1{S^2}}, \wt o\of{\frac{n}{mS^2}}},\ \ \
\eta_z\leq\wt o\of{\frac 1{mS^2}}
\end{equation}
and that at least one of the following three conditions holds:
\begin{equation} \label{eq:w_case}
m\leq{\wt o\of{nS^{1/4}}}, \ \ \ \frac{\eta_z}{\eta_w} \leq \wt
o\of{\frac Sm},
\end{equation}
\begin{equation} \label{eq:wz_case}
m\leq{\wt o\of{nS^{1/4}}}, \ \ \ \frac{\eta_z}{\eta_w} \geq \wt
\omega\of{\frac{m^2}{nS}},
\end{equation}
\begin{equation} \label{eq:z_case}
\frac{\eta_z}{\eta_w} \geq \wt
\omega\of{1+\frac mn}.
\end{equation}
Then with probability $\geq 1-(nS)^{-\omega(1)}$ w.r.t. the choice of
$X,W_0,z_0$ as above the following holds. For almost all {\rm (}in the Lebesque
sense{\rm )} $y\in\mathbb R^m$
the gradient descent {\rm\eqref{eq:gradient_descent}} avoids singular points and if additionally
$$
||y||\leq \wt O(m^{1/2}S^{1/2})
$$
then it converges to a global minimum.
\end{theorem}

Before embarking on the proof of this theorem, let us make a few remarks, in
addition to those that were already made in Section \ref{sec:contributions}.

\begin{remark}
The choice of labels is completely adversarial (they are only required to be
of ``right'' order), and the adversary is also allowed to see the
initialization, not only the data. This aligns well with the prominent
experiments in \cite{ZBH*} strongly suggesting that the choice of labels
should {\em not} be a defining factor in convergence.
\end{remark}

\begin{remark}
As in all previous work, the error probability can be decreased to
exponential by examining the proofs in Appendices \ref{app:properties_easy},
\ref{app:ntk_initialization}.
\end{remark}

\begin{remark} \label{rem:two_rates}
Since $m\leq S$, the bound in \eqref{eq:z_case} implies the second bound in
\eqref{eq:wz_case}. Thus, we always have either $m\leq{\wt o\of{nS^{1/4}}}$
or $\frac{\eta_z}{\eta_w} \geq \wt \omega\of{\frac{m^2}{nS}}$ and, in fact,
the refinement \eqref{eq:z_case} of this dichotomy will not be needed until
Appendix \ref{app:between_steps}.
\end{remark}

\section{Proof of the main result}

As noted before, we defer the simple proof that the gradient descent avoids
singular points a.e. to Appendix \ref{app:caveat} and for now simply assume
that $y\in\mathbb R^m$ is chosen in such a way that this is true, and such
that $||y||\leq \wt O(m^{1/2}S^{1/2})$.

We begin the proof of Theorem \ref{thm:main} with fixing some useful notation and reminding some basic facts.

First of all, it will be convenient to extend the trajectory \eqref{eq:gradient_descent} to
continuous time in the piecewise linear manner. That is, for any $t\geq 0$
we set
\begin{eqnarray*}
W_t &\df& W_{\lfloor t\rfloor} -\eta_w\{t\}\nabla^w\ell(\theta_{\lfloor t \rfloor});\\
z_t &\df& z_{\lfloor t\rfloor} -\eta_z\{t\}\nabla^z\ell(\theta_{\lfloor t \rfloor}).
\end{eqnarray*}

Since $\theta_\tau$ is regular for all $\tau\in\mathbb N$, the interval
$[\tau,\tau+1]$ may contain only finitely many $t$ for which $\theta_t$ is
singular. Hence those $t$ can and will be ignored
in our estimates based on integration.

We abbreviate $F(W_t), f(\theta_t), e(\theta_t)$ etc. to $F_t,f_t,e_t$
respectively. Whenever $\theta_t$ is regular, we let
$$
A_t\df \sigma'(W_tX)\ \ \ \text{(the activation matrix)}.
$$
If $\theta_t$ is singular, we let $A_t\df A_{t+0}$ which makes all our
constructs {\em upper} semi-continuous.

Next, let
$$
B_t\df \text{diag}(z_t)A_t\ \ \ \text{(the weighted activation matrix)},
$$
then $(B_t\ast X, F_t)$ is the transposed Jacobian $J_f(\theta_t)$ and hence
\begin{eqnarray*}
\nabla^w\ell(\theta_t) &=& (B_t\ast X)e_t;\\
\nabla^z\ell(\theta_t) &=& F_te_t.
\end{eqnarray*}
We also let
\begin{eqnarray*}
H_t &\df& (B_t\ast X)^\top (B_t\ast X) = (X^\top X)\circ (B_t^\top B_t),\\
G_t &\df& F_t^\top F_t;
\end{eqnarray*}
these are (for integer $t$) the two components of the NTK matrix.

\medskip
We will frame the rest of the proof according to the paradigm known in
combinatorics as {\em quasirandomness} (see \cite{CGW} for graphs and
\cite{quasirandomness} for arbitrary combinatorial objects). The idea is to
split a logically elaborated argument involving random objects into two
totally independent parts. At the first stage we accumulate a list of
so-called ``quasirandom properties'', that is {\em completely deterministic}
facts that hold for our random objects with overwhelming probability. At the
second stage, the argument proceeds completely deterministically, on the base
of these properties only. This allows us to avoid unwanted serious
complications, or even sheer mistakes, caused by the fact that quantifiers
and randomization do not get along well in one argument (see e.g. \cite{cliques}
where we adopted this approach). In the context of
deep learning, having a list of properties sufficient for convergence in a convenient form
might have another benefit: it may help to facilitate a discussion of what
{\em exactly} differentiates the data and initialization occurring in
practical problems from random synthetic data.

We would like to finish this brief discourse with acknowledging that this
approach is well known in the area under various names like ``assumptions'',
``meta-theorems'' (see e.g. \cite{OyS}) etc. The main difference is perhaps
that we do it a bit more systematically and on a larger scale.

\subsection{Quasirandom properties} \label{sec:properties}
We claim that with our choice of $X,W_0,z_0$ (Section \ref{sec:data}),
the following are satisfied with probability $\geq 1-(nS)^{-\omega(1)}$. The
proofs are deferred to Appendix \ref{app:properties_easy} (straightforward
proofs) and Appendix \ref{app:ntk_initialization} (not so straightforward).

\subsubsection{Properties of data}

\bigskip

\noindent {\bf Almost orthogonality.}
\begin{equation} \label{eq:almost_orthogonality}
\max_{j\neq j'\in [m]}\langle X^j, X^{j'}\rangle \leq \wt O(n^{1/2}).
\end{equation}

\smallskip
In the following property, as well as \eqref{eq:good}, the constants assumed
in the $\wt O$ notation do not depend on $k$ and $j,R$ respectively.

\noindent {\bf Uniform bounds on norm.}
\begin{equation} \label{eq:X_norm}
\forall k\in [m] \of{\max_{J\in {[m]\choose k}}||X^J||\leq \wt O\of{1+\frac{k^{1/2}}{n^{1/2}}}}.
\end{equation}

\smallskip
In the following property, as well as in \eqref{eq:z_large}, the bounds in
the existential quantifiers are completely explicit, see the proofs in
Appendix \ref{app:properties_easy}.

\noindent {\bf Dual bound on the minimum singular value.}

\begin{equation} \label{eq:X_dual}
\exists n^\ast\leq \wt O(n) \of{\min_{J\in {[m]\choose n^\ast}}\sigma_{\min}\of{(X^J)^\top}
\geq \frac nm}.
\end{equation}

\bigskip

\subsubsection{Properties of initialization}

\bigskip

\noindent {\bf Rows of $W_0$ are large.}
\begin{equation} \label{eq:large_rows}
\min_{\nu\in [S]} ||(W_0)_\nu|| \geq \wt\Omega(n^{1/2}).
\end{equation}

\medskip The following two properties are entirely obvious under either
Rademacher or Gaussian initialization of $z_0$.

\smallskip
\noindent{\bf Entries of $\theta_0$ are small.}
\begin{equation} \label{eq:z_entries}
||W_0||_\infty,\ ||z_0||_\infty \leq \wt O(1).
\end{equation}

\smallskip
\noindent{\bf $z_0$ contains sufficiently many large entries.}
\begin{equation} \label{eq:z_large}
\exists \zeta_0\geq \wt\Omega(1) \of{|\set{\nu\in [S]}{|(z_0)_\nu|\geq\zeta_0}|
\geq \wt\Omega(S)}.
\end{equation}

\bigskip

\subsubsection{Smooth properties}

\bigskip

\noindent{\bf Regularity.}
\begin{equation} \label{eq:regular}
\theta_0\ \text{is regular.}
\end{equation}

\smallskip
\noindent{\bf Data is almost orthogonal to initialization.}
\begin{equation} \label{eq:w0x}
||W_0X||_\infty \leq \wt O(1).
\end{equation}

\smallskip
\noindent{\bf Right order of the output at initialization.}
\begin{equation} \label{eq:f0}
||f_0||_\infty \leq \wt O(S^{1/2}).
\end{equation}

\smallskip
\noindent{\bf ``Good behavior'' for any data instance.}
\begin{equation} \label{eq:good}
\forall j\in [m] \forall R\geq 0 \of{|\set{\nu\in [S]}{|(W_0X)_{\nu j}|\leq R}|\leq
\wt O(SR+1)}.
\end{equation}

\bigskip

\subsubsection{NTK properties at initialization}

\bigskip

\noindent {\bf NTK: second layer.}
\begin{equation} \label{eq:ntk2}
\lambda_{\min}(G_0)\geq \Omega(S).
\end{equation}

\smallskip
In order to formulate our last (and the most difficult) property, let us fix
$\zeta_0\geq\wt\Omega(1)$ as in \eqref{eq:z_large}, and set
$$
\Gamma_0\df \set{\nu\in [S]}{|(z_0)_\nu|\geq \zeta_0};
$$
thus, $|\Gamma_0|\geq \wt\Omega(S)$.

\noindent {\bf NTK: first layer.}
\begin{equation} \label{eq:ntk1}
\begin{split}
m\leq \widetilde o(nS^{1/2}) &\Longrightarrow \exists S^\ast \geq
\wt\Omega\of{\frac{n^2S}{n^2+m}}\\ &\of{\min_{\Gamma\in {\Gamma_0\choose
|\Gamma_0|- S^\ast}}\lambda_{\min}\of{((A_0)_\Gamma\ast X)^\top
((A_0)_\Gamma\ast X)}\geq \Omega(S)}.
\end{split}
\end{equation}

\begin{remark}
We have not been able to remove the annoying term $m$ in the denominator of
the bound on $S^\ast$; we believe it can be done with a better analysis than
ours.
\end{remark}

\subsection{Deterministic part} \label{sec:det_part}

According to the plan outlined above, from now on we are assuming that
$X,\theta_0$ are chosen arbitrarily in such a way that all properties in
Section \ref{sec:properties} are satisfied. The labels $y\in\mathbb R^m$ are
also arbitrary, as long as $||y||\leq \wt O(m^{1/2}S^{1/2})$ and the gradient
descent avoids singular points. The learning rates $\eta_w, \eta_z$ are
assumed to satisfy \eqref{eq:absolute_bounds} and one of \eqref{eq:w_case},
\eqref{eq:wz_case}, \eqref{eq:z_case}. All the way until Appendix
\ref{app:between_steps}, however, we will only need that either $m\leq \wt
o(nS^{1/4})$ or $\frac{\eta_z}{\eta_w} \geq \wt\omega\of{\frac{m^2}{nS}}$.

\begin{remark} \label{rem:dichotomy}
Before we begin, let us make one technical remark. The
definition \eqref{eq:asymptotic_simple} of the assumption $m\leq \wt
o(nS^{1/4})$ can be equivalently re-written as
\begin{equation} \label{eq:asymptotic}
\lim_{k\to\infty} \frac{\log\of{\frac{n^{(k)}
(S^{(k)})^{1/4}}{m^{(k)}}}}{\log\log\of{n^{(k)} S^{(k)}}}=\infty.
\end{equation}
Likewise, $m\geq\wt\Omega(nS^{1/4})$ means that the quantity in
\eqref{eq:asymptotic} is bounded. Since from every infinite sequence we can
extract an infinite subsequence for which one of the two is true, we can
freely assume w.l.o.g. that either $m\leq \wt o(nS^{1/4})$ or
$m\geq\wt\Omega(nS^{1/4})$ holds. The same dichotomy applies to
$\frac{\eta_z}{\eta_w} \geq \wt\omega\of{\frac{m^2}{nS}}$ vs.
$\frac{\eta_z}{\eta_w} \leq \wt O\of{\frac{m^2}{nS}}$.
\end{remark}

\bigskip
We are going to prove the following upper bounds on the distance travelled in
the parameter space:
\begin{eqnarray}
\label{eq:bound1} m\leq \wt o\of{nS^{1/4}} &\Longrightarrow& \sup_{T>0}||W_T-W_0||_{\text{\tiny F}}<
\wt O(m^{1/2});\\
\label{eq:bound2} m\leq \wt o\of{nS^{1/4}} &\Longrightarrow& \sup_{T>0}||z_T-z_0||_{\text{\tiny F}}<
\wt O\of{\frac{\eta_z^{1/2}}{\eta_w^{1/2}}m^{1/2}};\\
\label{eq:bound3} \frac{\eta_z}{\eta_w} \geq
\wt\omega\of{\frac{m^2}{nS}} &\Longrightarrow& \sup_{T>0}||W_T-W_0||_{\text{\tiny F}}<
\wt  O\of{\frac{\eta_w^{1/2}}{\eta_z^{1/2}}m^{1/2}};\\
\label{eq:bound4}  \frac{\eta_z}{\eta_w} \geq
\wt\omega\of{\frac{m^2}{nS}}  &\Longrightarrow& \sup_{T>0}||z_T-z_0||_{\text{\tiny F}}<
\wt O\of{m^{1/2}}.
\end{eqnarray}

With a slight abuse of notation\footnote{To be completely impeccable, we
should first fix the constants assumed in the right-hand sides.}, we prove
\eqref{eq:bound1}-\eqref{eq:bound4} by induction on $T$; that is, we fix
$T>0$ and assume that these bounds hold for all $t<T$. We then claim that for
all $t<T$ we also have
\begin{eqnarray}
\label{eq:ntkh} m\leq \wt o\of{nS^{1/4}} &\Longrightarrow& \lambda_{\min}(H_t)\geq
\wt\Omega(S);\\
\label{eq:ntkg} \frac{\eta_z}{\eta_w} \geq
\wt\omega\of{\frac{m^2}{nS}}  &\Longrightarrow& \lambda_{\min}(G_t)\geq
 \wt\Omega(S).
\end{eqnarray}

Let us prove \eqref{eq:ntkg} since it is easier. Given \eqref{eq:ntk2}, it is
sufficient to show that
\begin{equation} \label{eq:F_bound}
||F_t-F_0|| \leq \wt o(S^{1/2}).
\end{equation}

For that we perform the calculation
\begin{equation} \label{eq:frobenius_calculation}
\begin{split}
||F_t-F_0|| &\leq ||F_t-F_0||_{\text{\tiny F}} = ||\sigma(W_tX)-\sigma(W_0X)||_{\text{\tiny F}}\\&\leq ||(W_t-W_0)X||_{\text{\tiny F}}
\stackrel{\eqref{eq:handy}}{\leq}{||W_t-W_0||_{\text{\tiny F}}\cdot ||X||},
\end{split}
\end{equation}
where the third inequality hold since $\sigma$ is 1-Lipschitz. We now check
that \eqref{eq:frobenius_calculation} indeed implies \eqref{eq:F_bound}.

First,
\begin{equation} \label{eq:W_universal}
||W_t-W_0||_{\text{\tiny F}}\stackrel{\eqref{eq:bound3}}{\leq} \wt
O\of{\frac{\eta_w^{1/2}}{\eta_z^{1/2}}m^{1/2}} \leq \wt
o\of{\frac{n^{1/2}S^{1/2}}{m^{1/2}}},
\end{equation}
where for the second inequality we used the assumption in \eqref{eq:ntkg}.

Next, as noted in Remark \ref{rem:dichotomy}, we can assume w.l.o.g. that
either $m\geq \wt \Omega\of{nS^{1/4}}$ or $m\leq \wt o\of{nS^{1/4}}$.

In the first case, the bound \eqref{eq:X_norm} for $k=m$ simplifies to
$||X||\leq \wt O\of{\frac{m^{1/2}}{n^{1/2}}}$. This, along with \eqref{eq:W_universal}, implies \eqref{eq:F_bound}.

If, on the other hand, $m\leq \wt o\of{nS^{1/4}}$, we can also employ \eqref{eq:bound1} and refine \eqref{eq:W_universal} to
$$
||W_t-W_0||_{\text{\tiny F}}\leq  \min\of{\wt O(m^{1/2}),\ \wt
o\of{\frac{n^{1/2}S^{1/2}}{m^{1/2}}}} \leq \min\of{\wt o(S^{1/2}),\ \wt
o\of{\frac{n^{1/2}S^{1/2}}{m^{1/2}}}}.
$$
Multiplying this with the full version $||X||\leq \wt O\of{1+\frac{m^{1/2}}{n^{1/2}}}$ of \eqref{eq:X_norm} again gives us \eqref{eq:F_bound}. Thus, the proof of \eqref{eq:F_bound} and hence also of \eqref{eq:ntkg} is completed.

\medskip
The proof of \eqref{eq:ntkh} is more elaborate and is deferred to Appendix \ref{app:ntk}.

\medskip
The dynamics of the gradient descent in the error space is ruled by the differential equation
$$
\stackrel .\ell_t =\langle\nabla\ell(\theta_t), \stackrel .\theta_t\rangle,
$$
where
$$
\stackrel .\theta_t = - \of{\eta_w\nabla^w\ell(\theta_{\lfloor t\rfloor}),
\eta_z\nabla^z\ell(\theta_{\lfloor t\rfloor})}.
$$
Hence
$$
\frac{d}{dt}||e_t|| = \frac{\frac d{dt}||e_t||^2}{2||e_t||} =\frac{\stackrel .\ell_t}{||e_t||} =  \frac{\langle\nabla\ell(\theta_t),
\stackrel .\theta_t\rangle}{||e_t||} =
\frac{\langle\nabla\ell(\theta_t), \stackrel .\theta_{\lfloor t\rfloor}\rangle}{||e_t||}.
$$
Let
$$
\delta_t\df \nabla\ell(\theta_t)-\nabla\ell(\theta_{\lfloor t\rfloor})
$$
(this measures how much the gradient may change during one step). We now have
$$
\langle\nabla\ell(\theta_t), \stackrel .\theta_{\lfloor t\rfloor}\rangle =
\langle\nabla\ell(\theta_{\lfloor t\rfloor}),
\stackrel .\theta_{\lfloor t\rfloor}\rangle + \langle\delta_t,
\stackrel .\theta_{\lfloor t\rfloor}\rangle =
-\eta_w||\nabla^w\ell(\theta_{\lfloor t \rfloor})||^2 - \eta_z||\nabla^z\ell(\theta_{\lfloor t \rfloor})||^2
+ \langle\delta_t,
\stackrel .\theta_{\lfloor t\rfloor}\rangle.
$$

In Appendices \ref{app:between_steps}, \ref{app:invariant} we will prove that
\begin{equation} \label{eq:between_steps}
\absvalue{\langle\delta_t,
\stackrel .\theta_{\lfloor t\rfloor}\rangle} < \frac 12
\langle\nabla\ell(\theta_{\lfloor t\rfloor}), \stackrel .\theta_{\lfloor t\rfloor}\rangle
\end{equation}
(the proof for the $z$-part is pretty straightforward, given the absolute
bounds \eqref{eq:absolute_bounds}, but will require a new idea for the
$W$-part due to the discontinuity of $\nabla^w\ell$).

Once we have that, we know that $||e_t||$ is decreasing or, more
specifically,
\begin{equation} \label{eq:decay}
\frac{d}{dt}||e_t|| \leq \frac{\langle\nabla\ell(\theta_{\lfloor t\rfloor}),
{\stackrel .\theta}_{\lfloor t\rfloor}\rangle}{2||e_t||} \leq \frac{\langle\nabla\ell(\theta_{\lfloor t\rfloor}),
{\stackrel .\theta}_{\lfloor t\rfloor}\rangle}{2||e_{\lfloor t\rfloor}||},
\end{equation}
where the last inequality holds since $||e_t||$ is decreasing.

Note that $||f_0||\leq \wt O(m^{1/2}S^{1/2})$ by \eqref{eq:f0} and,
therefore, $||e_0||\leq \wt O(m^{1/2}S^{1/2})$ due to our assumption on $y$.
Integrating \eqref{eq:decay} from 0 to $T$ gives us
$$
\int_0^T \frac{\langle\nabla\ell(\theta_{\lfloor t\rfloor}), -
{\stackrel .\theta}_{\lfloor t\rfloor}\rangle}{||e_{\lfloor t\rfloor}||}dt\leq 2\int_0^T
\of{-\frac{d}{dt}||e_t||}dt = 2(||e_0||-||e_T||)\leq \wt O(m^{1/2}S^{1/2}),
$$
where the equality holds since $||e_t||$ is decreasing.

Using Cauchy-Schwartz, we now estimate
\begin{equation} \label{eq:w_pre}
\begin{split}
||W_T-W_0||_{\text{\tiny F}} &\leq \eta_w\int_0^T ||\nabla^w\ell(\theta_{\lfloor t\rfloor})|| dt \\& \leq
\eta_w\of{\int_0^T\frac{||\nabla^w\ell(\theta_{\lfloor t\rfloor})||^2}{||e_{\lfloor t\rfloor}||}}^{1/2}\cdot \of{\int_0^T ||e_{\lfloor t\rfloor}||dt}^{1/2}
\\&\leq \eta_w
\of{\int_0^T \frac{\langle\nabla\ell(\theta_{\lfloor t\rfloor}), -
\stackrel .\theta_t\rangle}{\eta_w||e_{\lfloor t\rfloor}||}dt}^{1/2} \cdot \of{\int_0^T ||e_{\lfloor t\rfloor}||dt}^{1/2}\\& \leq \wt O\of{\eta_w^{1/2}m^{1/4}S^{1/4}\cdot \of{\int_0^T ||e_{\lfloor t\rfloor}||dt}^{1/2}}
\end{split}
\end{equation}
and, likewise,
\begin{equation} \label{eq:z_pre}
||z_T-z_0|| \leq \wt O\of{\eta_z^{1/2}m^{1/4}S^{1/4}\cdot \of{\int_0^T ||e_{\lfloor t\rfloor}||dt}^{1/2}}.
\end{equation}

In order to estimate the last remaining term $\int_0^T ||e_{\lfloor
t\rfloor}||dt$, let us first assume that $m\leq\wt o(nS^{1/4})$. In that case
we have \eqref{eq:ntkh} and then we can continue \eqref{eq:decay} as follows:
\begin{equation*}
\begin{split}
\frac d{dt}||e_t|| &\leq -\wt\Omega\of{\frac{\langle\nabla\ell(\theta_{\lfloor t\rfloor}), -
\stackrel .\theta_{\lfloor t\rfloor}\rangle}{||e_{\lfloor t\rfloor}||}} \leq
-\wt\Omega\of{\eta_w\frac{||\nabla^w\ell(\theta_{\lfloor t\rfloor})||^2}{||e_{\lfloor t\rfloor}||}}
\\&\leq - \wt\Omega(\eta_w\lambda_{\min}(H_{\lfloor t\rfloor})||e_{\lfloor t \rfloor}||) \leq
-\wt\Omega(\eta_wS||e_{\lfloor t\rfloor}||).
\end{split}
\end{equation*}
This (sub)differential equation solves to
\begin{equation} \label{eq:e_solution}
\begin{split}
m\leq \wt o(nS^{1/4}) \Longrightarrow ||e_t|| &\leq ||e_0||\cdot \exp(-\wt\Omega(\eta_wtS)) \\&\leq \wt O\of{m^{1/2}S^{1/2}\exp(-\wt\Omega(\eta_wtS)}.
\end{split}
\end{equation}

Integrating from 0 to $T$ gives us
\begin{equation} \label{eq:e_w}
m\leq \wt o(nS^{1/4}) \Longrightarrow \int_0^T ||e_{\lfloor t\rfloor}||dt\leq \wt O\of{\frac{m^{1/2}}{\eta_w S^{1/2}}}
\end{equation}
and, likewise (but applying \eqref{eq:ntkg} this time),
\begin{eqnarray}
\label{eq:ez_solution} \frac{\eta_z}{\eta_w} \geq \wt\omega\of{\frac{m^2}{nS}} &\Longrightarrow& ||e_t|| \leq \wt O\of{m^{1/2}S^{1/2}\exp(-\wt\Omega(\eta_ztS)};\\
\label{eq:e_z} \frac{\eta_z}{\eta_w} \geq \wt\omega\of{\frac{m^2}{nS}} &\Longrightarrow& \int_0^T ||e_{\lfloor t\rfloor}||dt\leq \wt O\of{\frac{m^{1/2}}{\eta_z S^{1/2}}}.
\end{eqnarray}
Plugging \eqref{eq:e_w}, \eqref{eq:e_z} into \eqref{eq:w_pre}, \eqref{eq:z_pre} gives us all four inequalities \eqref{eq:bound1}-\eqref{eq:bound4}. This completes their proof by joint induction on $t$.

\smallskip
In particular, in the first case \eqref{eq:w_case} we have \eqref{eq:e_solution}, in the third case \eqref{eq:z_case} we have \eqref{eq:ez_solution} (and in the second case \eqref{eq:wz_case} we have both). This ``completes''\footnote{The quotation marks indicate the fact that all serious work is delegated to Appendix.} the proof of Theorem \ref{thm:main}, with exponential speed of convergence.

\section{Experiments} \label{sec:experiments}

Since our paper does not claim any direct relevance to practical data, we have confined our experiments to synthetic data generated precisely as in Section \ref{sec:data}, with $\zeta\in_R \{\pm 1\}$. In most experiments, the vector $y$ of label data was chosen from $\mathcal N(0,S)^m$. We also performed a few experiments with other natural choices:
\begin{description}
\item[low spectrum] $e_0=f_0-y$ is the eigenvector of the matrix $H_0$
    corresponding to the smallest eigenvalue and scaled in such a way that
    $||e_0|| = m^{1/2}S^{1/2}$.

\item[high spectrum] $e_0=n^{1/2}S^{1/2}(X^\top X^1)$. This is a ``natural choice'' for which $||H_0e_0||$ is large.

\item[local] $e_0= \begin{pmatrix}
                     m^{1/2}S^{1/2} \\
                     0 \\
                     \vdots \\
                     0
                   \end{pmatrix}$. The initial error is completely concentrated on the first data instance (and is huge).
\end{description}
We considered only the case $\eta_z=0$, i.e., training the first layer. All our experiments were equipped with the ``safety valve'': they terminated once the situation $||e_\tau||>||e_{\tau-1}||$ was encountered. A computation was declared successful and stopped once $||e_\tau||<10^{-3}$. We remark that due to our choice of normalization, the initial value was roughly of order $10^2-10^4$ so our stopping condition corresponds to the loss reduction by a factor of $10^{10}-10^{14}$.

Like any other theoretical paper, ours is plugged with asymptotic notation (we would like to note in passing that attempts to replace it with explicit constants are usually not very instructive or even readable). The original purpose of our experiments was rather limited: verify that our theoretical findings are still reasonably relevant for relatively small values of parameters. What they have showed instead is that the crucial quantity $\lambda_{\min}(H_t)$ still behaves smoothly and nicely well beyond what could be optimistically called the ``NTK regime'', almost all the way up to the representability barrier $m\sim nS$. In simpler words, there should be a significant room for improving theoretical results on convergence with methods yet to be developed. In retrospect, this is not surprising at all: the bounds like the first inequality in \eqref{eq:w_pre} are hopelessly pessimistic. But it is nice to have an experimental encouragement.

\medskip
Let us now be more specific. In all experiments the dimension was set to
$n:=100$ and, unless otherwise noted, the learning rate was
$\eta_w:=10^{-3}$. We did not try to optimize on the latter since this is not
the main focus of the paper. But several sporadic experiments showed that for
smaller values of $S,m$ it can be significantly increased.

We kept track of the following control quantities (here $T$ is the stopping time):
\begin{itemize}
\item $\kappa_H \df \frac{\lambda_{\min}(H_T)}{\lambda_{\min}(H_0)}$.
    Computing this quantity is computationally costly so we did not track
    the evolution of $\lambda_{\min}(H_\tau)$ systematically. But a
    spike-like behavior ($\lambda_{\min}(H_\tau)$ goes down and then up
    again) was never observed in several sporadic experiments in which it was tracked.

\item $|D|$, where $D\subseteq [S]\times [m]$ is the set of all pairs $(\nu, j)$ that change their activation at least once during training;

\item $||W_T-W_0||_{\text{\tiny F}}$. Same remark as in the first item
    applies (except that it is not very costly).
\end{itemize}

\smallskip
In the main series of our experiments, the vector $y\in\mathbb R^m$ was
chosen from $\mathcal N(0,S)^m$, and we considered four values of $S$:
$S=100,\ 200,\ 500,\ 1000$. The number of input data $m$ ranged from 100 to
1000, with step $\frac S{10}$. For every triple $(n,S,m)$ we performed 10
experiments, with fresh values of $(X, y, W_0,z_0)$ each time. {\em All these
experiments resulted in convergence as defined above} but for $S\in
\{500,1000\},\ m\geq 900$ we had to decrease the learning rate to
$\eta_w=5\cdot 10^{-4}$ when $S=500$ and to $\eta_w=2\cdot 10^{-4}$ when
$S=1000$.

\begin{figure}[ht]
\begin{center}
\epsfig{file=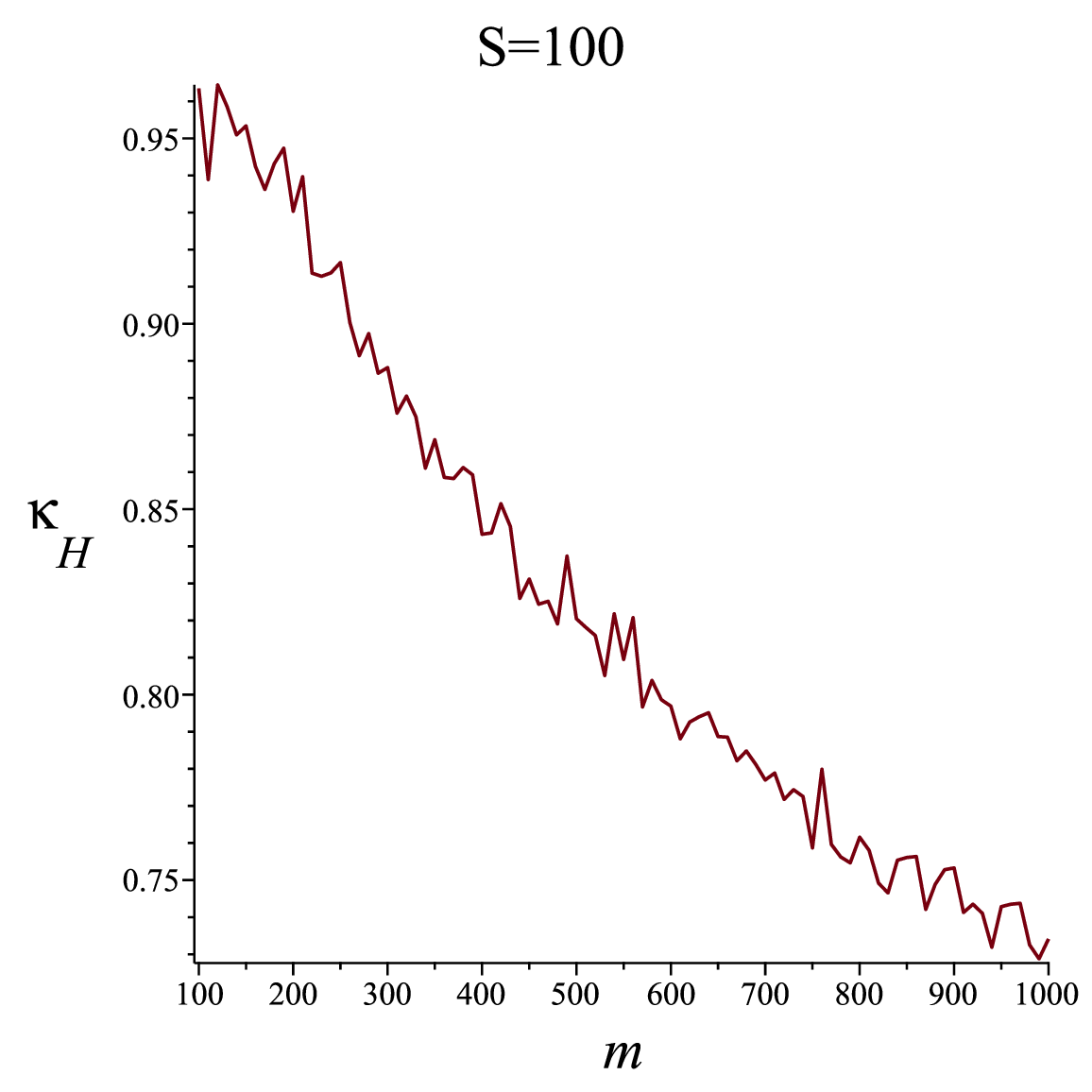,width=5cm}\ \hspace{3cm} \epsfig{file=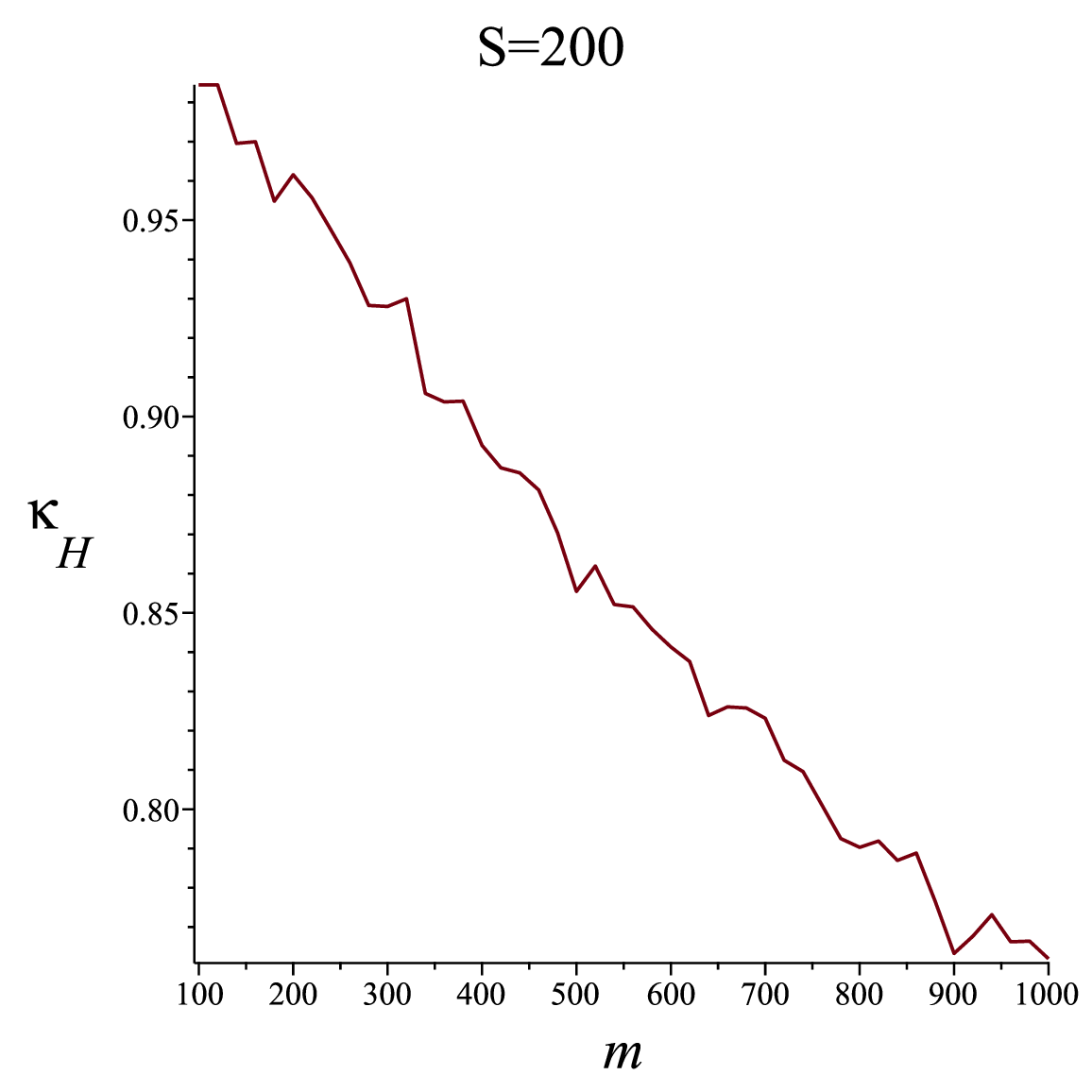,width=5cm}
\\ \vspace{1cm} \epsfig{file=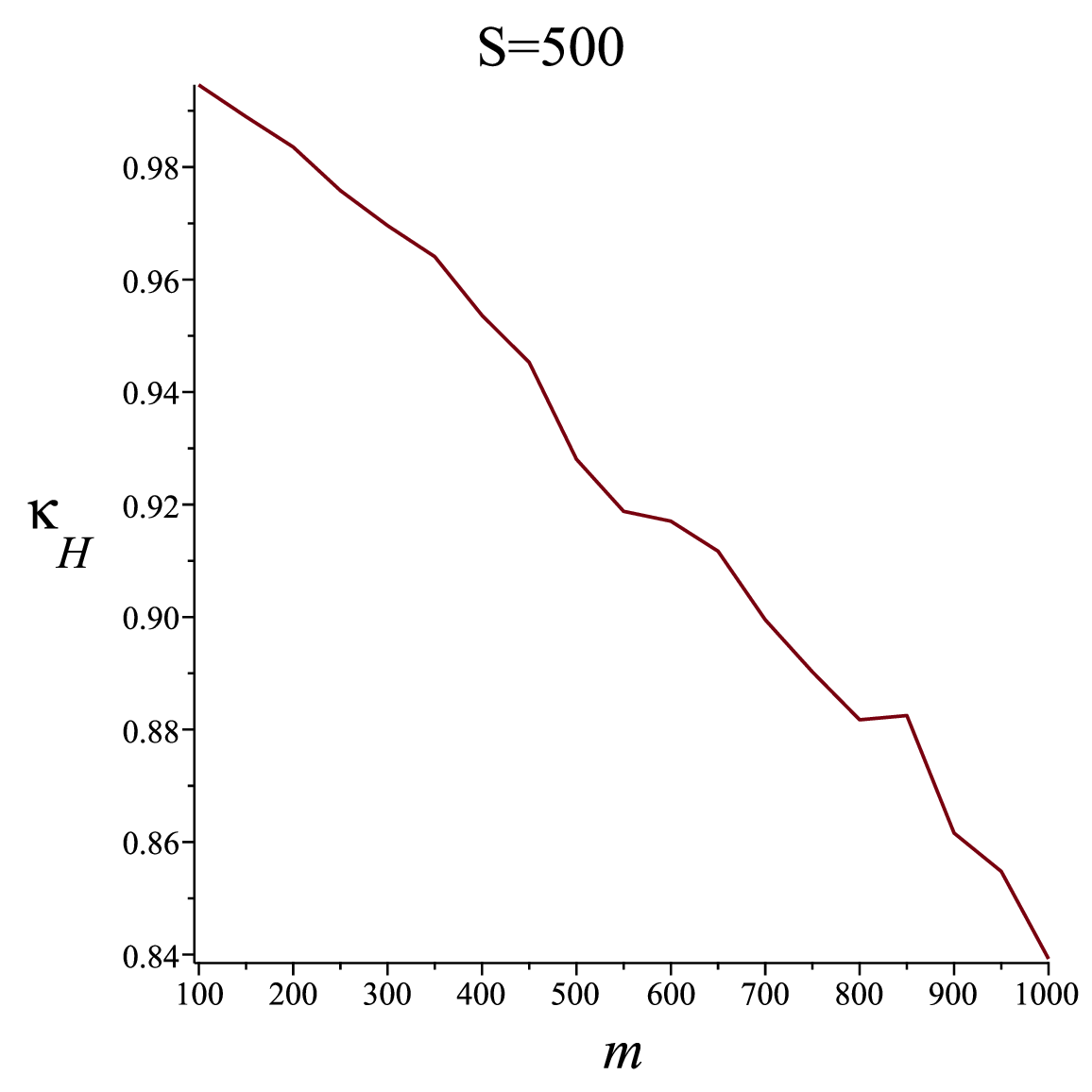,width=5cm}\ \hspace{3cm} \epsfig{file=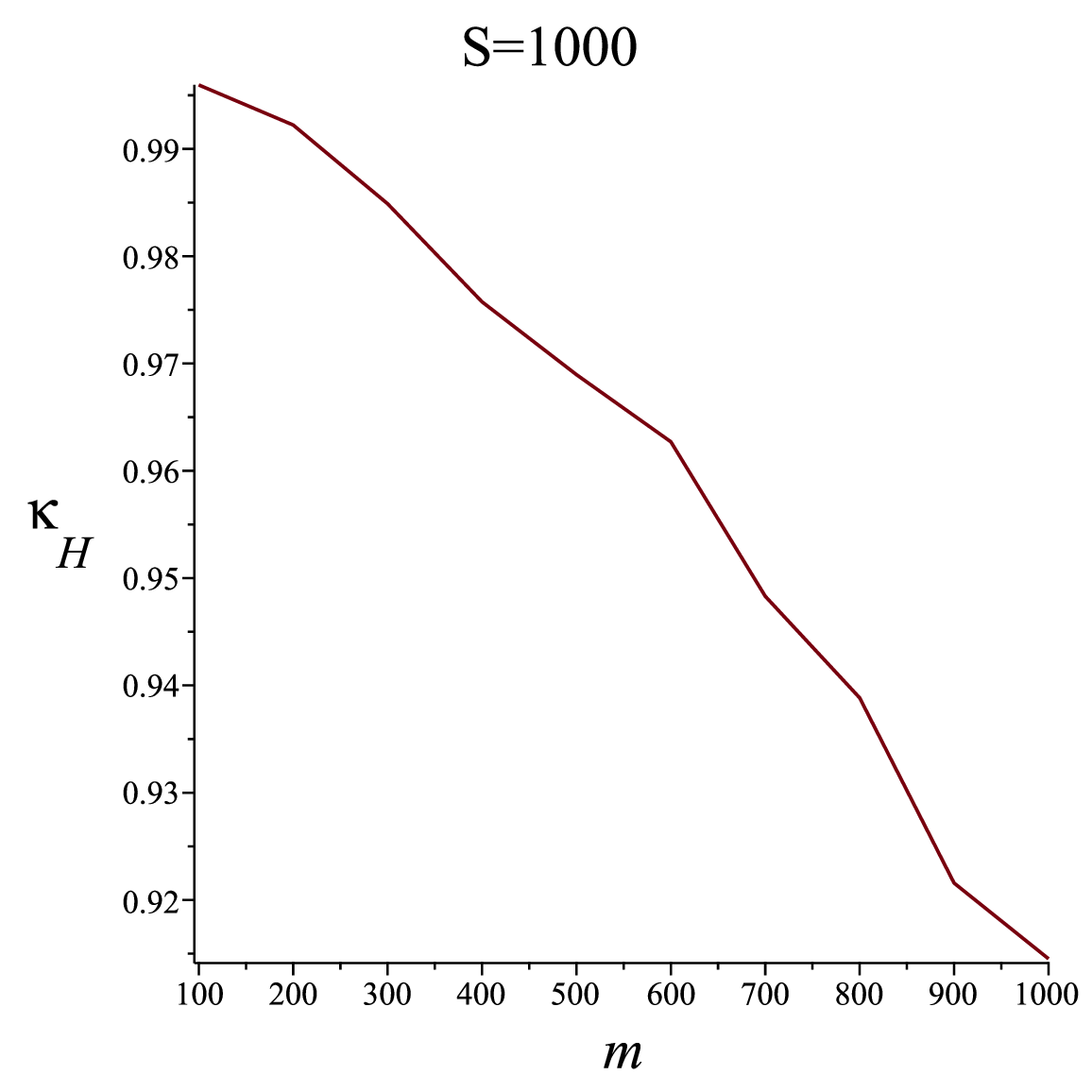,width=5cm}
\caption{\small The dependence of $\kappa_H$ on $m$. \label{fig:kappa}}
\end{center}
\end{figure}

\begin{figure}[ht]
\begin{center}
\epsfig{file=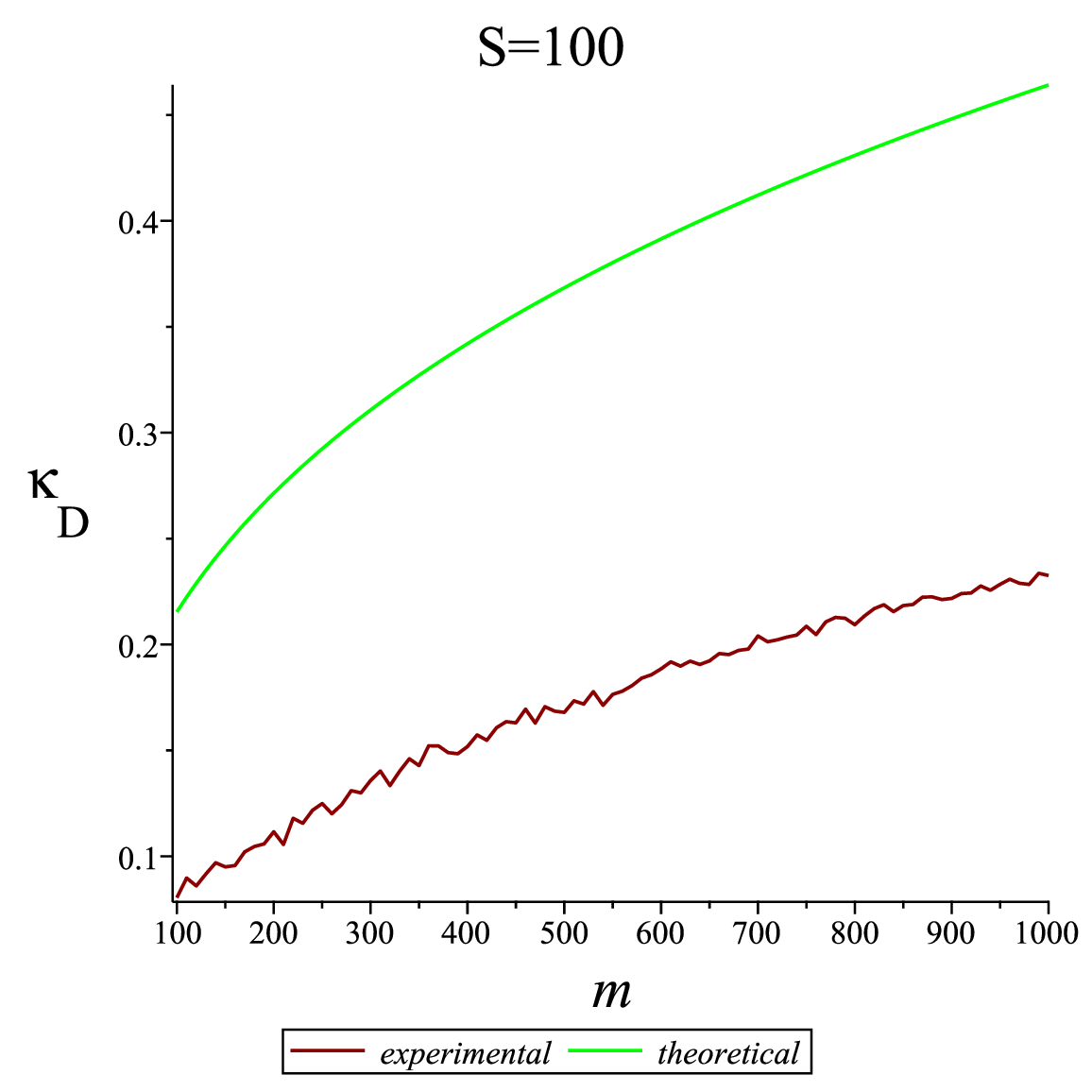,width=5cm}\ \hspace{3cm} \epsfig{file=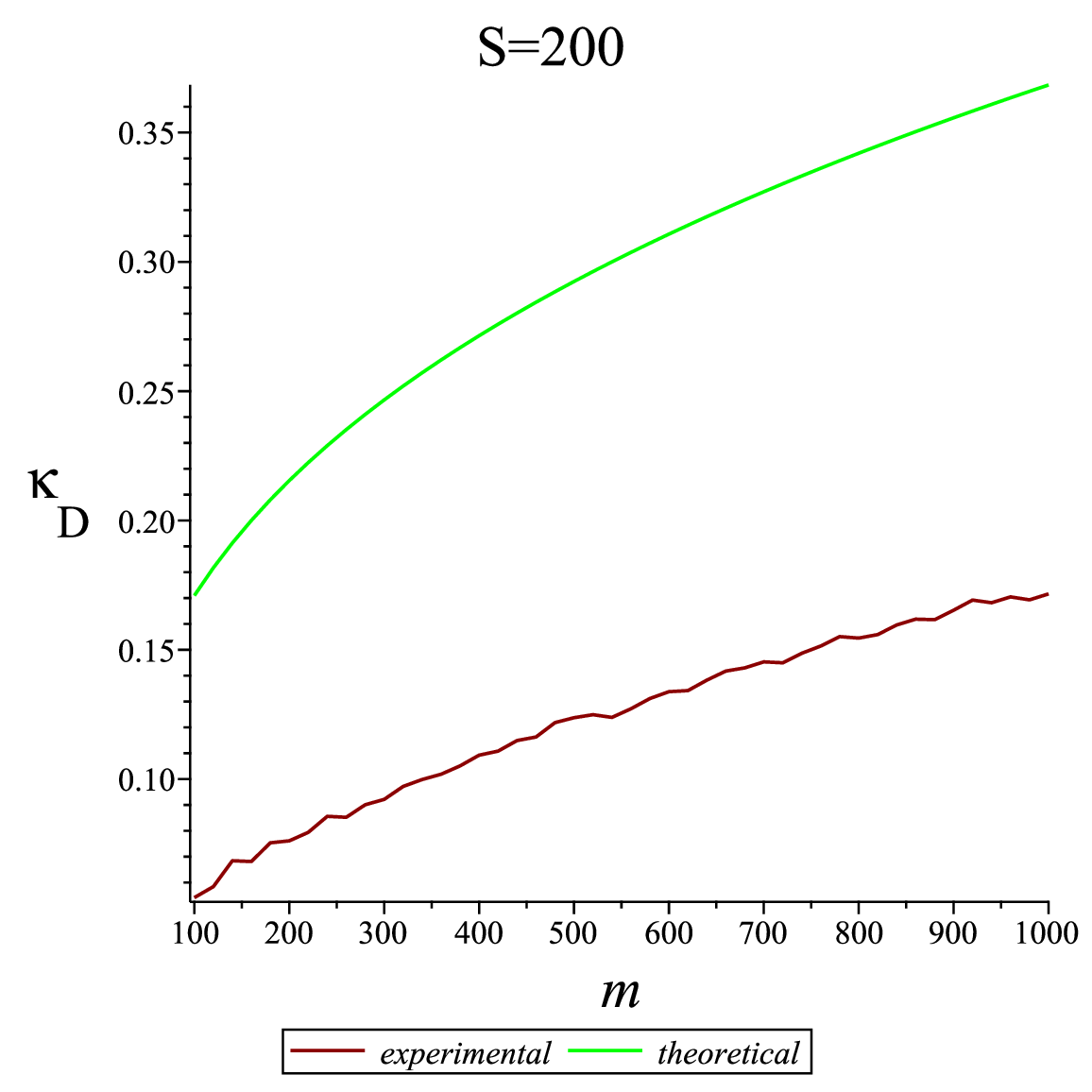,width=5cm}
\\ \vspace{1cm} \epsfig{file=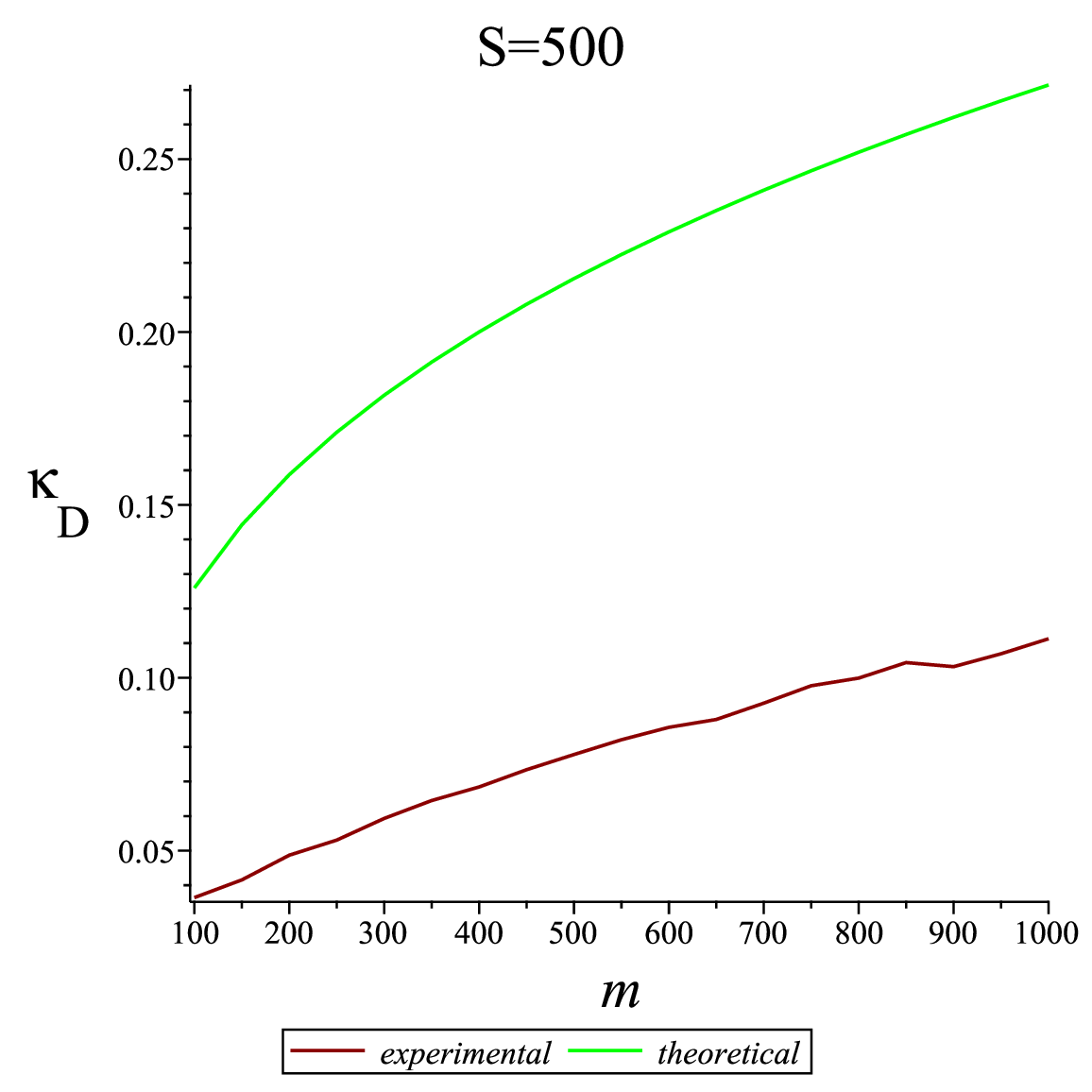,width=5cm}\ \hspace{3cm} \epsfig{file=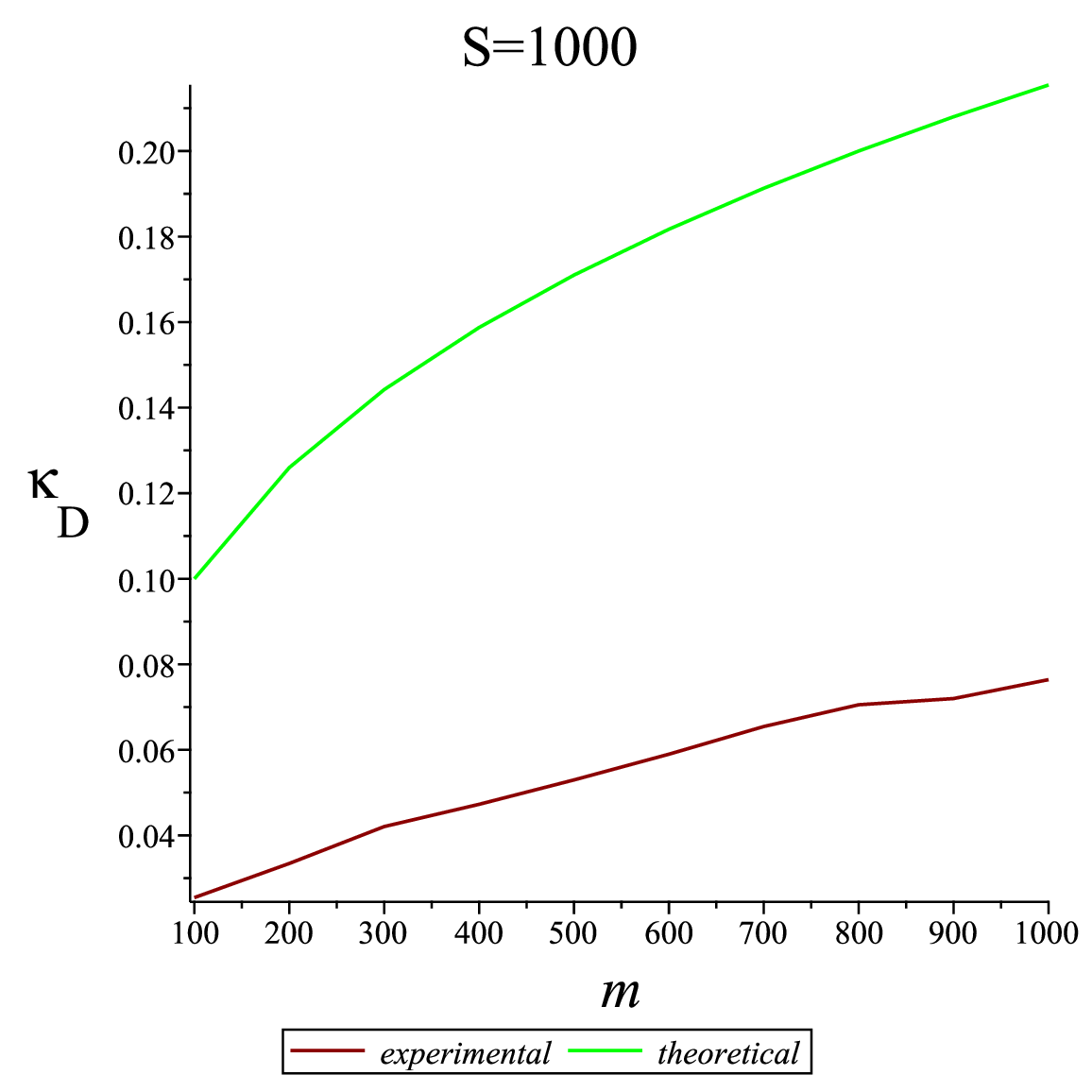,width=5cm}
\caption{\small $\kappa_D\df \frac{|D|}{|mS|}$ is the {\em density} of changes. The
green line shows the theoretical prediction $\of{\frac{m}{nS}}^{1/3}$ obtained by
dividing  \eqref{eq:bound_on_s} by $S$. \label{fig:D}}
\end{center}
\end{figure}

\begin{figure}[ht]
\begin{center}
\epsfig{file=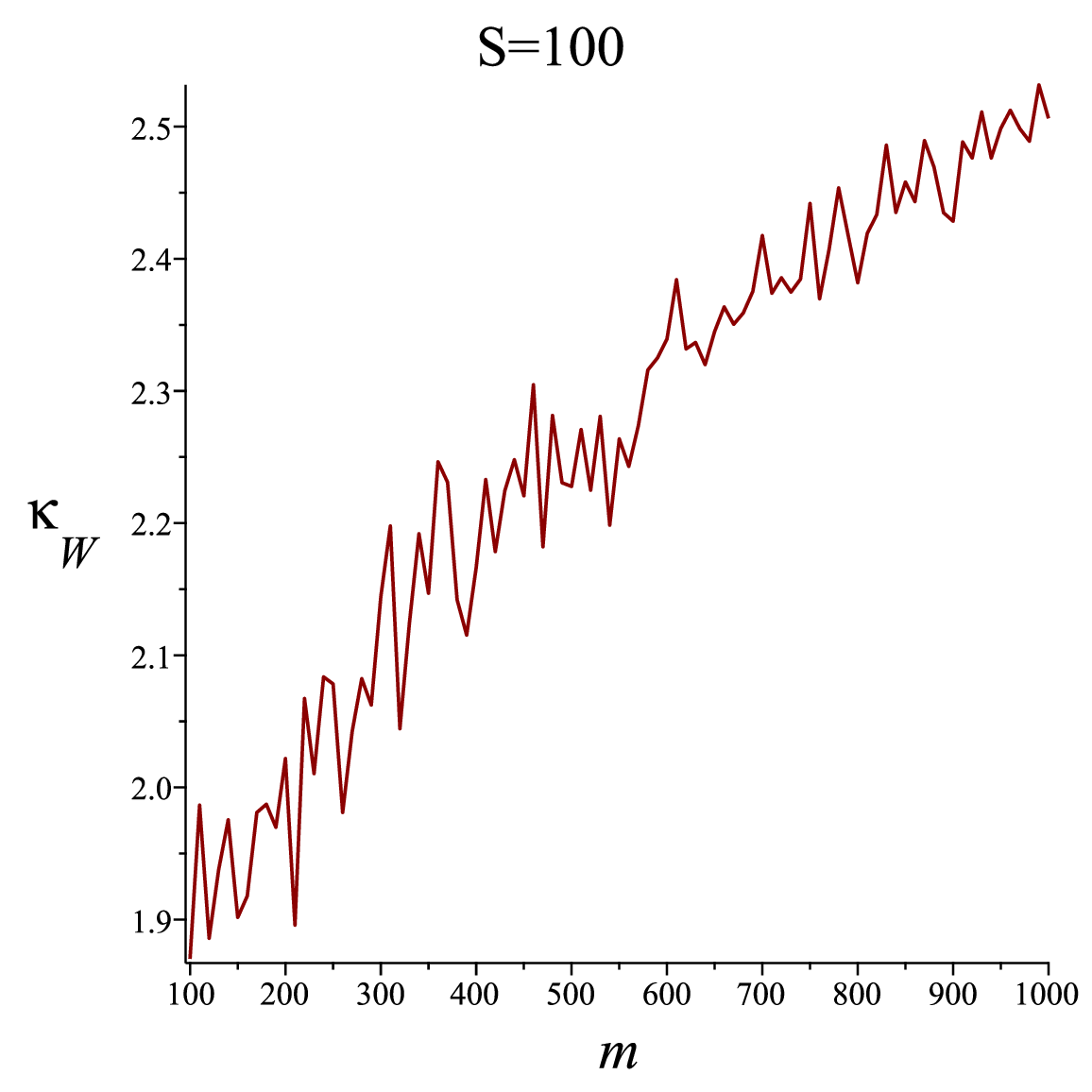,width=5cm}\ \hspace{3cm} \epsfig{file=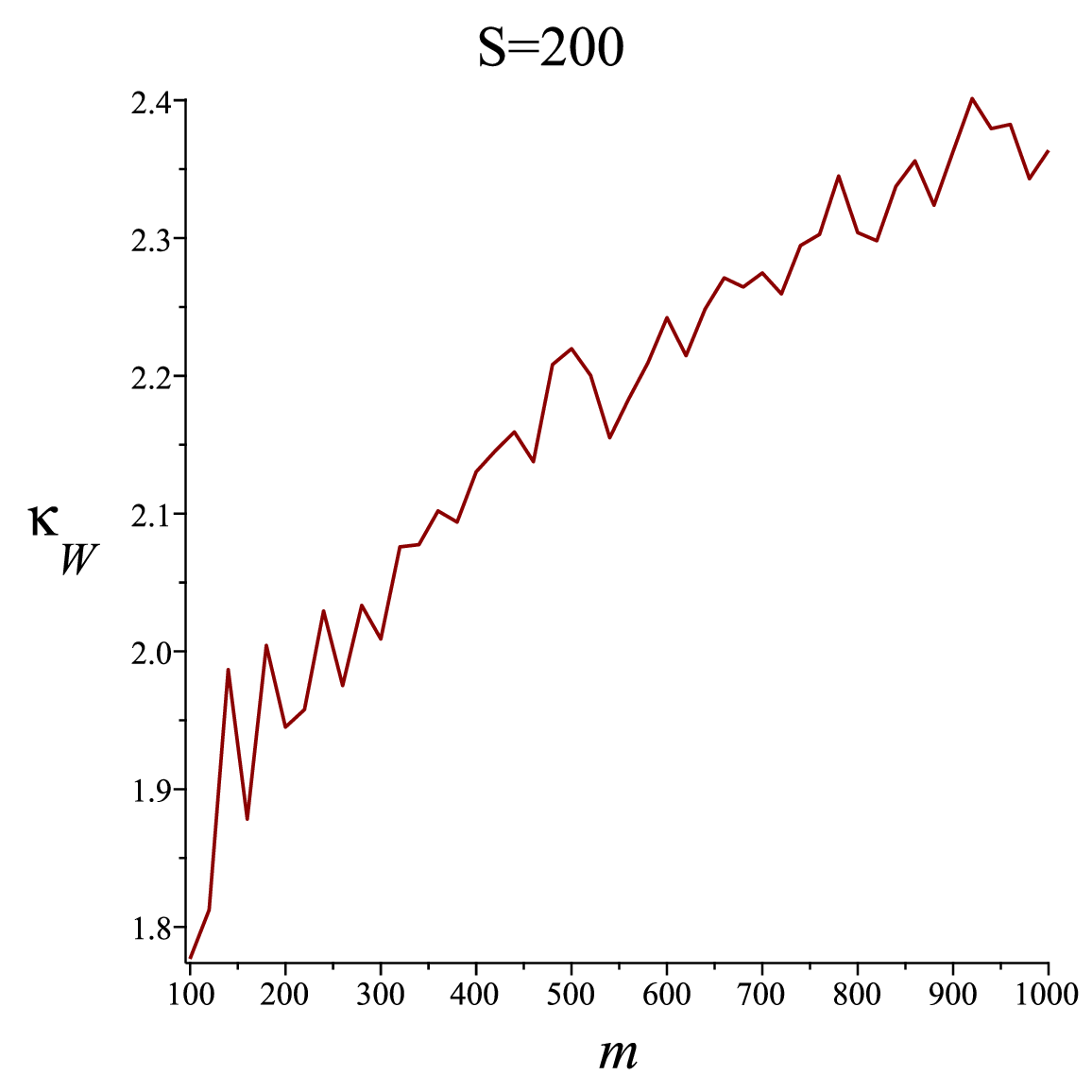,width=5cm}
\\ \vspace{1cm} \epsfig{file=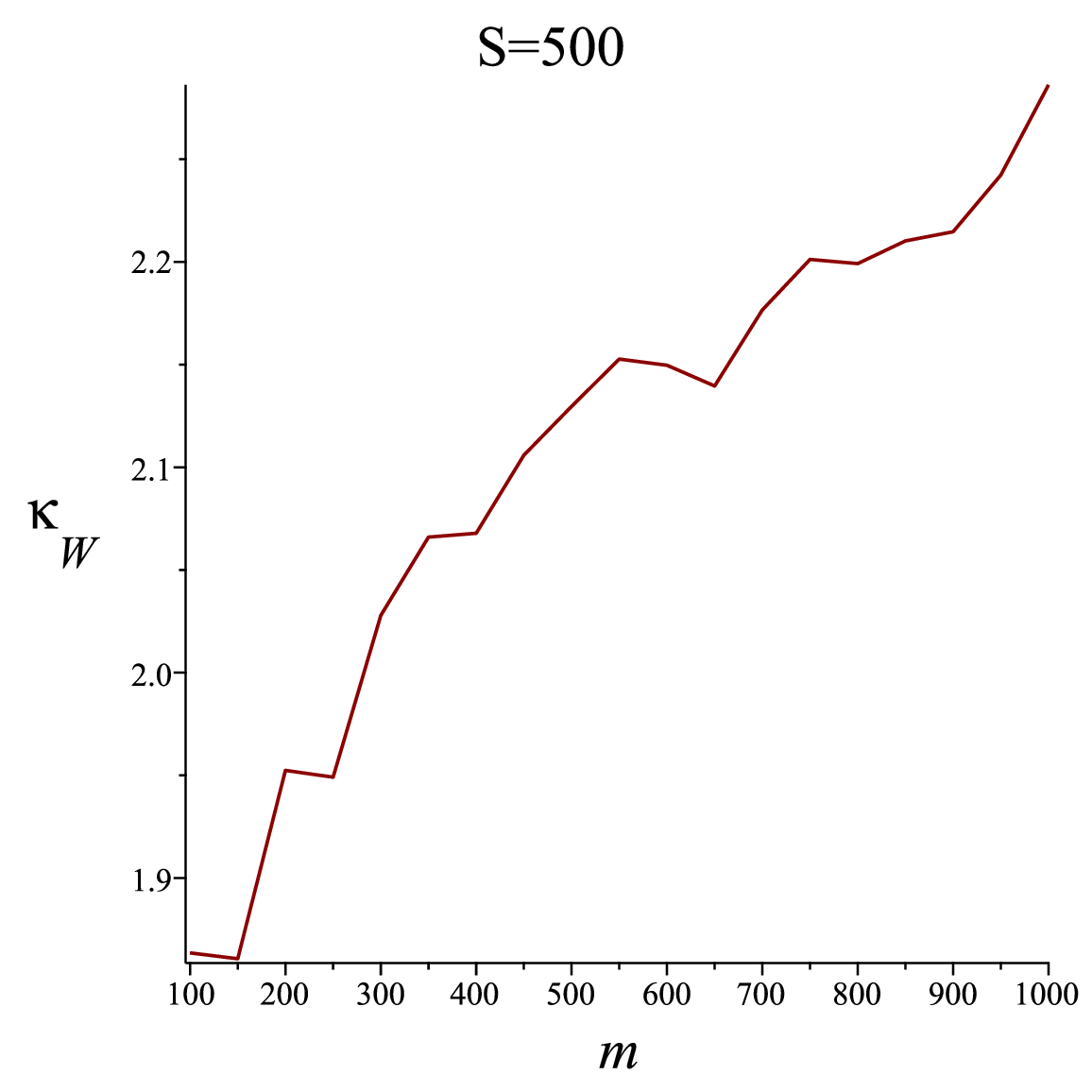,width=5cm}\ \hspace{3cm} \epsfig{file=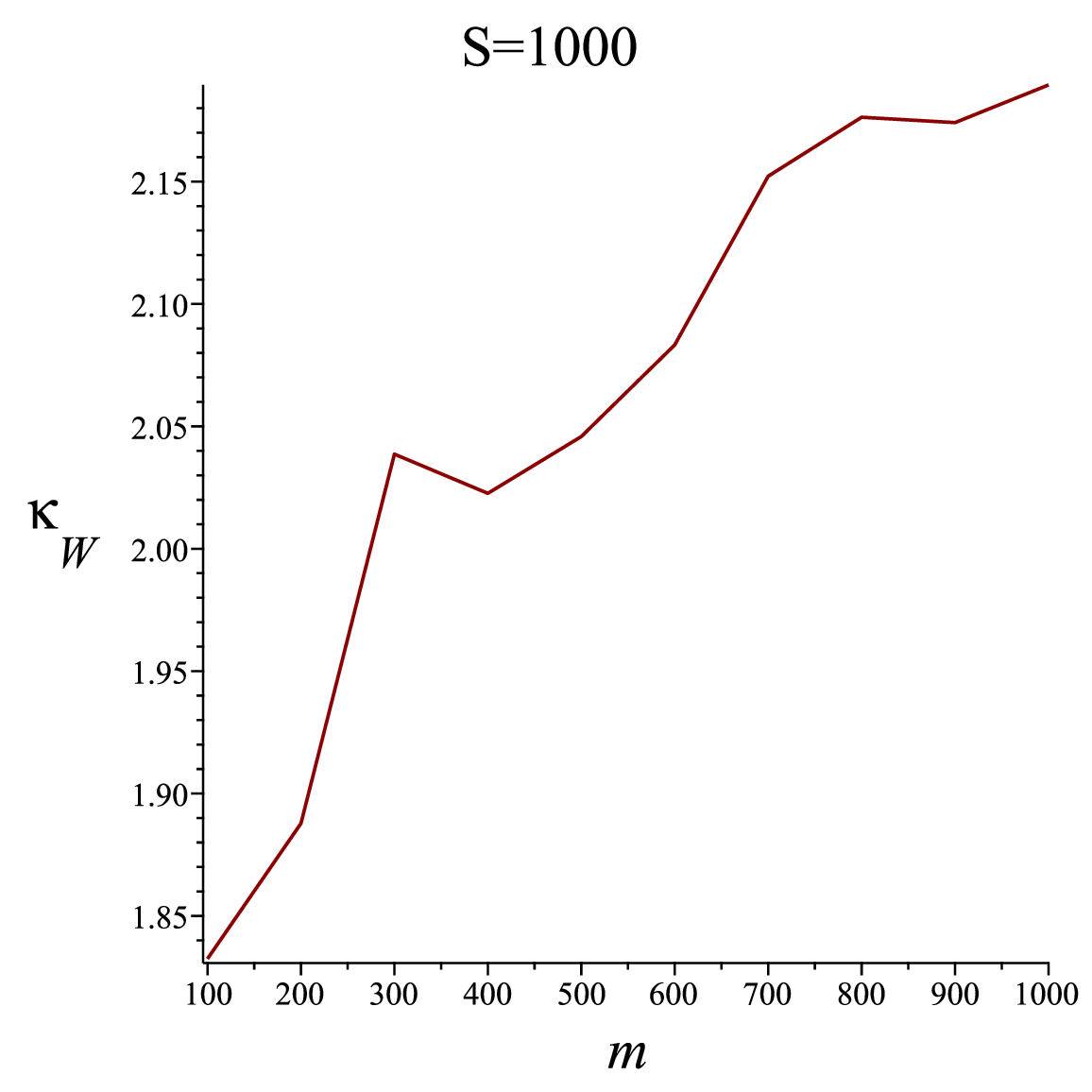,width=5cm}
\caption{\small $\kappa_W\df \frac{||W_T-W_0||_{\text{\tiny F}}}{m^{1/2}}$ (cf.
\eqref{eq:bound1}) \label{fig:W}}
\end{center}
\end{figure}

The ranges for our control parameters are partially\footnote{To make it
comprehensible, we confine ourselves to $m=100,\ 200,\,\ldots, 1000$ for all
four values of $S$. Our main purpose for presenting results in the tabular
form is to demonstrate that there is sufficiently sharp concentration in 10
experiments performed for every pair $(S,m)$.} reported in Table
\ref{tab:table1}, and their average values are depicted on Figures
\ref{fig:kappa}, \ref{fig:D} and \ref{fig:W}; for further comments and
explanations see the respective captions.

\smallskip
In less systematic way, for $S=100$ we also tried really large values (note
that $nS=10000$) $m=1000,\ 2000,\ 3000,\ 4000, \ 5000$, with 10,\ 10,\ 6,\ 3
and 1 experiments, respectively. All of them have converged, Table \ref{tab:table2}
tabulates the results of these experiments; note that it quite
smoothly extends the part of Table \ref{tab:table1} for $S=100$.

We also ran a few experiments (again, for $S=100$) with other choices of the
labels $y$ mentioned above. No significant discrepancies have been found with
the high spectrum case and lower spectrum case.
\begin{table}[ht]
\hspace{-2.5cm}
\begin{tabular}{|c|c|c|c|c|c|} \hline
$\boldsymbol{S}$ & $\boldsymbol{m}$ & $\boldsymbol{T}$  & $\boldsymbol{\kappa_H}$ & $\boldsymbol{|D|}$ & $\boldsymbol{||W_T-W_0||_F}$  \\
\hline\hline

100 & 100 & 468-624; \bf{528} & 0.89- 1.00; \bf{0.96} & 657-1024; \bf{805} & 15.91-22.86; \bf{18.70} \\ \hline
100 & 200 & 664-772; \bf{699} & 0.88- 0.96; \bf{0.93} & 2091-2482; \bf{2232} & 27.10-31.74; \bf{28.59} \\ \hline
100 & 300 & 825-905; \bf{861} & 0.84- 0.91; \bf{0.89} & 3801-4449; \bf{4074} & 34.81-39.73; \bf{37.14} \\ \hline
100 & 400 & 931-1054; \bf{991} & 0.82- 0.88; \bf{0.84} & 5763-6952; \bf{6072} & 40.81-48.48; \bf{43.33} \\ \hline
100 & 500 & 1091-1176; \bf{1129} & 0.80- 0.84; \bf{0.82} & 8133-8802; \bf{8397} & 47.43-51.65; \bf{49.81} \\ \hline
100 & 600 & 1216-1388; \bf{1282} & 0.74- 0.84; \bf{0.80} & 10858-12893; \bf{11311} & 55.23-63.03; \bf{57.30} \\ \hline
100 & 700 & 1362-1492; \bf{1418} & 0.76- 0.82; \bf{0.78} & 13496-15722; \bf{14282} & 60.32-68.55; \bf{63.96} \\ \hline
100 & 800 & 1489-1623; \bf{1547} & 0.72- 0.78; \bf{0.76} & 15974-17881; \bf{16747} & 61.82-70.55; \bf{67.37} \\ \hline
100 & 900 & 1591-1816; \bf{1685} & 0.71- 0.80; \bf{0.75} & 19126-21581; \bf{19958} & 68.92-78.38; \bf{72.85} \\ \hline
100 & 1000 & 1784-1915; \bf{1840} & 0.71- 0.75; \bf{0.73} & 22411-24654; \bf{23255} & 76.52-81.95; \bf{79.26} \\ \hline    \hline
200 & 100 & 235-264; \bf{245} & 0.95- 1.00; \bf{0.98} & 888-1451; \bf{1084} & 14.68-22.89; \bf{17.77} \\ \hline
200 & 200 & 306-330; \bf{314} & 0.92- 0.98; \bf{0.96} & 2785-3308; \bf{3045} & 26.09-29.24; \bf{27.51} \\ \hline
200 & 300 & 362-384; \bf{368} & 0.91- 0.94; \bf{0.93} & 5163-5971; \bf{5530} & 32.50-38.23; \bf{34.80} \\ \hline
200 & 400 & 406-441; \bf{425} & 0.87- 0.92; \bf{0.89} & 8116-9286; \bf{8740} & 39.43-45.76; \bf{42.61} \\ \hline
200 & 500 & 446-505; \bf{480} & 0.83- 0.90; \bf{0.86} & 11173-13078; \bf{12373} & 43.74-52.91; \bf{49.63} \\ \hline
200 & 600 & 508-538; \bf{524} & 0.82- 0.86; \bf{0.84} & 15004-17063; \bf{16056} & 51.39-58.14; \bf{54.92} \\ \hline
200 & 700 & 547-590; \bf{570} & 0.81- 0.84; \bf{0.82} & 19367-21210; \bf{20345} & 57.03-63.41; \bf{60.18} \\ \hline
200 & 800 & 610-654; \bf{632} & 0.77- 0.82; \bf{0.79} & 23235-25827; \bf{24723} & 61.05-67.69; \bf{65.17} \\ \hline
200 & 900 & 674-707; \bf{691} & 0.75- 0.77; \bf{0.76} & 28437-31031; \bf{29753} & 68.31-74.35; \bf{70.89} \\ \hline
200 & 1000 & 706-785; \bf{728} & 0.72- 0.78; \bf{0.76} & 32606-36192; \bf{34320} & 70.12-80.08; \bf{74.74} \\ \hline   \hline
500 & 100 & 91-98; \bf{95} & 0.98- 1.00; \bf{0.99} & 1654-2065; \bf{1822} & 16.23-20.71; \bf{18.64} \\ \hline
500 & 200 & 114-122; \bf{117} & 0.98- 0.99; \bf{0.98} & 4356-5386; \bf{4869} & 24.75-30.71; \bf{27.61} \\ \hline
500 & 300 & 129-137; \bf{133} & 0.95- 0.98; \bf{0.97} & 8453-9347; \bf{8898} & 32.96-37.27; \bf{35.12} \\ \hline
500 & 400 & 144-150; \bf{148} & 0.94- 0.97; \bf{0.95} & 13192-14614; \bf{13684} & 39.68-43.53; \bf{41.36} \\ \hline
500 & 500 & 159-166; \bf{163} & 0.92- 0.94; \bf{0.93} & 18258-20345; \bf{19450} & 44.44-49.85; \bf{47.62} \\ \hline
500 & 600 & 172-178; \bf{175} & 0.90- 0.93; \bf{0.92} & 24445-26661; \bf{25699} & 48.71-55.22; \bf{52.66} \\ \hline
500 & 700 & 182-196; \bf{187} & 0.87- 0.93; \bf{0.90} & 30777-35985; \bf{32429} & 54.55-65.46; \bf{57.59} \\ \hline
500 & 800 & 193-209; \bf{201} & 0.86- 0.91; \bf{0.88} & 36625-42393; \bf{39962} & 57.16-66.52; \bf{62.20} \\ \hline
500 & 900 & 427-459; \bf{437} & 0.84- 0.88; \bf{0.86} & 44111-48943; \bf{46445} & 62.97-70.43; \bf{66.44} \\ \hline
500 & 1000 & 463-478; \bf{470} & 0.83- 0.85; \bf{0.84} & 52602-58394; \bf{55638} & 68.06-75.92; \bf{72.29} \\ \hline      \hline
1000 & 100 & 94-101; \bf{96} & 0.99- 1.00; \bf{1.00} & 2123-3066; \bf{2546} & 15.78-22.89; \bf{18.33} \\ \hline
1000 & 200 & 112-119; \bf{115} & 0.98- 1.00; \bf{0.99} & 5831-7786; \bf{6688} & 22.94-31.31; \bf{26.70} \\ \hline
1000 & 300 & 125-131; \bf{130} & 0.98- 1.00; \bf{0.98} & 11194-14368; \bf{12620} & 31.81-40.68; \bf{35.31} \\ \hline
1000 & 400 & 137-146; \bf{142} & 0.96- 0.99; \bf{0.98} & 17651-19942; \bf{18908} & 37.78-43.71; \bf{40.45} \\ \hline
1000 & 500 & 147-156; \bf{152} & 0.95- 0.99; \bf{0.97} & 25194-27879; \bf{26485} & 43.62-49.31; \bf{45.75} \\ \hline
1000 & 600 & 158-166; \bf{162} & 0.95- 0.98; \bf{0.96} & 33262-37519; \bf{35391} & 48.37-53.29; \bf{51.03} \\ \hline
1000 & 700 & 168-180; \bf{174} & 0.92- 0.98; \bf{0.95} & 43723-48665; \bf{45819} & 54.62-59.54; \bf{56.94} \\ \hline
1000 & 800 & 180-189; \bf{185} & 0.92- 0.95; \bf{0.94} & 53662-60491; \bf{56434} & 57.42-66.78; \bf{61.55} \\ \hline
1000 & 900 & 487-503; \bf{498} & 0.91- 0.94; \bf{0.92} & 62774-67471; \bf{64791} & 63.01-68.03; \bf{65.22} \\ \hline
1000 & 1000 & 511-532; \bf{520} & 0.90- 0.93; \bf{0.91} & 74727-78658; \bf{76410} & 67.13-70.88; \bf{69.24} \\ \hline
\end{tabular}
\caption{Average values are shown in bold. \label{tab:table1}}
\end{table}

\begin{table}[ht]
\hspace{-2.5cm}
\begin{tabular}{|c|c|c|c|c|} \hline
$\boldsymbol{m}$ & $\boldsymbol{T}$  & $\boldsymbol{\kappa_H}$ & $\boldsymbol{|D|}$ & $\boldsymbol{||W_T-W_0||_F}$  \\
\hline\hline

1000 & $1739- 1948; \mathbf{1859}$ & $0.7- 0.76; \mathbf{0.73}$ & $21835- 24793; \mathbf{23321}$ & $74.62- 83.61; \mathbf{79.73}$\\

\hline

2000 & $3363- 3731; \mathbf{3581}$ & $0.62- 0.68; \mathbf{0.64}$ & $57838- 61459; \mathbf{59157}$ & $125.67- 136.35; \mathbf{131.03}$\\

\hline

3000 & $6482- 7088; \mathbf{6689}$ & $0.53- 0.57; \mathbf{0.56}$ & $99676- 102597; \mathbf{100896}$ & $186- 192.5; \mathbf{189.2}$\\

\hline

4000 & $11558- 12199; \mathbf{11984}$ & $0.48- 0.5; \mathbf{0.49}$ & $146339- 146489; \mathbf{146410}$ & $245- 255.3; \mathbf{251.8}$\\

\hline
5000  & \bf 28217 & \bf 0.36 & \bf 210147 & \bf 360.8  \\
\hline

\end{tabular}
\caption{\small $S=100$, $m$ is large. Average values are shown in bold. The values
for $m=1000$ were recomputed anew.  \label{tab:table2}}
\end{table}

\clearpage

The situation appears more interesting in the local case, that is when the
initial error is placed on just one data point; in fact, this is the only
situation where we have discovered anything that looks like a sharp
transition. To report on the most striking experiment, for $n=S=100,\
m=1000,\ \eta_w=10^{-3}$ the gradient descent converged 19 times out of 20.
While for $m=2000$, even after decreasing $\eta_w$ to $10^{-4}$, it aborted
(according to the stopping rule $||e_\tau|| > ||e_{\tau-1}||$) in 10 (out of
10) attempts. In yet another experiment, decreasing the learning rate even
further did not help convergence. But since these effects seem to be
happening well beyond the range $m\ll S$ covered by our (and all previous)
paper, we have not tried to investigate this systematically but rather defer
this to future research.

\section{Conclusion}

In this paper we have improved on previously known convergence guarantees of
the plain gradient descent in the case of shallow networks. The real
significance of this whole line of work depends on the yet elusive
properties of data and labels that allow for the generalization. In the
optimistic scenario (from convergence to generalization) this research will
be eventually extended into the practical region and then it might turn out
to be helpful to explain why certain choices of labels allow us not only to
converge but also to generalize (again, cf. \cite{ZBH*}). In the other
scenario, generalization will be explained in such a way that the implied
convergence will clearly follow from this ad-hoc, label-dependent explanation. That
might render the whole line or research this paper belongs to obsolete.

But while the jury is out, let us briefly sketch what, in our view, are
worthy directions for future work assuming the first scenario.

First and foremost, it would be nice to be able to extend our knowledge for
depth 2 networks to networks of larger constant depth, without significantly
weakening the results.

It would be extremely nice (and probably very difficult, too) to prove
convergence outside of the NTK regime, i.e. find more intelligent ways of
controlling the learning trajectory than simply by the distance travelled. A
significant improvement along these lines would be to remove or relax the
assumption $m\ll S$ ubiquitously present in our and all previous work. The
invariant from \cite{ACH,DHL} that we will explore in Appendix
\ref{app:invariant} seems to be a paradigmatic example of the tools that are
to be developed for the purpose.

It would be interesting to pinpoint those ``quasirandom properties'' from
Section \ref{sec:properties} (as well as the previous work) that are most
problematic from the practical viewpoint. That might also give a good
indication of what might be promising directions to generalize known results.

Finally, many (if not most) practical applications do not fit the set-up in
Section \ref{sec:set-up}: they either deal with (multi)-classification
problems or with more sophisticated network architectures, like CNN or
Resent, or both. A great deal of interesting work was done in these
directions; some of it was cited in Section \ref{sec:intro}. But our
impression is that less effort has been invested into working out more
general united theory that would be less prone to this kind of changes in the
model. That, in our view,  is another interesting, and potentially more
accessible, goal.


\begin{thebibliography}{ADH{\etalchar{+}}19b}

\bibitem[ABR{\etalchar{+}}21]{cliques} A.~Atserias, I.~Bonacina, S.~Rezende,
    M.~Lauria, J.~{Nordstr\"om}, and
  A.~Razborov.
\newblock Clique is hard on average for regular resolution.
\newblock {\em Journal of the ACM}, 68(4), 2021.

\bibitem[ACH18]{ACH} Sanjeev Arora, Nadav Cohen, and Elad Hazan.
\newblock On the optimization of deep networks: Implicit acceleration by
  overparameterization.
\newblock In {\em Proceedings of the {\rm 35}th International Conference on
  Machine Learning {\rm (}ICML{\rm )}}, pages 244--253, 2018.

\bibitem[ADH{\etalchar{+}}19a]{ADH2*} Sanjeev Arora, Simon~S. Du, Wei Hu,
    Zhiyuan Li, Ruslan Salakhutdinov, and
  Ruosong Wang.
\newblock On exact computation with an infinitely wide neural net.
\newblock In {\em Proceedings of the {\rm 33}rd Conference on Neural
  Information Processing Systems {\rm (}NeurIPC{\rm )}}, pages 8139--8148,
  2019.

\bibitem[ADH{\etalchar{+}}19b]{ADH*} Sanjeev Arora, Simon~S. Du, Wei Hu,
    Zhiyuan Li, and Ruosong Wang.
\newblock Fine-grained analysis of optimization and generalization for
  overparameterized two-layer neural networks.
\newblock In {\em Proceedings of the {\rm 36}th International Conference on
  Machine Learning {\rm (}ICML{\rm )}}, pages 322--332, 2019.

\bibitem[AGNZ18]{AGN*} Sanjeev Arora, Rong Ge, Behnam Neyshabur, and
    Yi~Zhang.
\newblock Stronger generalization bounds for deep nets via a compression
  approach.
\newblock In {\em Proceedings of the {\rm 35}th International Conference on
  Machine Learning {\rm (}ICML{\rm )}}, pages 254--263, 2018.

\bibitem[ALS19]{ALS} Zeyuan Allen{-}Zhu, Yuanzhi Li, and Zhao Song.
\newblock A convergence theory for deep learning via over-parameterization.
\newblock In {\em Proceedings of the {\rm 36}th International Conference on
  Machine Learning {\rm (}ICML{\rm )}}, pages 242--252, 2019.

\bibitem[BAM22]{BAM} Simone Bombari, Mohammad~Hossein Amani, and Marco
    Mondelli.
\newblock Memorization and optimization in deep neural networks with minimum
  over-parameterization.
\newblock Technical Report 2205.10217 [stat.ML], arxiv e-print, 2022.

\bibitem[BFT17]{BFT} Peter~L. Bartlett, Dylan~J. Foster, and Matus Telgarsky.
\newblock Spectrally-normalized margin bounds for neural networks.
\newblock In {\em Proceedings of the {\rm 31}st Conference on Neural
  Information Processing Systems {\rm (}NeurIPC{\rm )}}, pages 6240--6249,
  2017.

\bibitem[BGMS18]{BGM*} Alon Brutzkus, Amir Globerson, Eran Malach, and Shai
    Shalev{-}Shwartz.
\newblock {SGD} learns over-parameterized networks that provably generalize on
  linearly separable data.
\newblock In {\em Proceedings of the {\rm 6}th International Conference on
  Learning Representations ({\rm (}ICLR{\rm )})}, 2018.

\bibitem[BHMM19]{BHM*} Mikhail Belkin, Daniel Hsu, Siyuan Ma, and Soumik
    Mandal.
\newblock Reconciling modern machine-learning practice and the classical
  bias-variance trade-off.
\newblock {\em Proceedings of the National Academy of Sciences},
  116(32):15849--15854, 2019.

\bibitem[CCZG21]{CCZ*} Zixiang Chen, Yuan Cao, Difan Zou, and Quanquan Gu.
\newblock How much over-parameterization is sufficient to learn deep {ReLU}
  networks?
\newblock In {\em Proceedings of the {\rm 9}th International Conference on
  Learning Representations ({\rm (}ICLR{\rm )})}, 2021.

\bibitem[CGW89]{CGW} F.~Chung, R.~Graham, and R.~Wilson.
\newblock Quasi-random graphs.
\newblock {\em Combinatorica}, 9:345--362, 1989.

\bibitem[COB19]{COB} L{\'{e}}na{\"{\i}}c Chizat, Edouard Oyallon, and
    Francis~R. Bach.
\newblock On lazy training in differentiable programming.
\newblock In {\em Proceedings of the {\rm 33}rd Conference on Neural
  Information Processing Systems {\rm (}NeurIPC{\rm )}}, pages 2933--2943,
  2019.

\bibitem[CR21]{quasirandomness} L.~Coregliano and A.~Razborov.
\newblock Natural quasirandom properties.
\newblock Technical Report 2012.11773 [math.CO], arxiv e-print, 2021.

\bibitem[DHL18]{DHL} Simon~S. Du, Wei Hu, and Jason~D. Lee.
\newblock Algorithmic regularization in learning deep homogeneous models:
  Layers are automatically balanced.
\newblock In {\em Proceedings of the {\rm 32}nd Conference on Neural
  Information Processing Systems {\rm (}NeurIPC{\rm )}}, pages 382--393, 2018.

\bibitem[DZPS19]{DZR*} Simon~S. Du, Xiyu Zhai, Barnab{\'{a}}s P{\'{o}}czos,
    and Aarti Singh.
\newblock Gradient descent provably optimizes over-parameterized neural
  networks.
\newblock In {\em Proceedings of the {\rm 7}th International Conference on
  Learning Representations ({\rm (}ICLR{\rm )})}, 2019.

\bibitem[FCB22]{FCB} Spencer Frei, Niladri~S. Chatterji, and Peter~L.
    Bartlett.
\newblock Benign overfitting without linearity: Neural network classifiers
  trained by gradient descent for noisy linear data.
\newblock In {\em Proceedings of the {\rm 35}th Conference on Learning Theory
  {\rm (}COLT{\rm )}}, pages 2668--2703, 2022.

\bibitem[HY20]{HuY} Jiaoyang Huang and Horng{-}Tzer Yau.
\newblock Dynamics of deep neural networks and neural tangent hierarchy.
\newblock In {\em Proceedings of the {\rm 37}th International Conference on
  Machine Learning {\rm (}ICML{\rm )}}, pages 4542--4551, 2020.

\bibitem[JT20]{JiT} Ziwei Ji and Matus Telgarsky.
\newblock Polylogarithmic width suffices for gradient descent to achieve
  arbitrarily small test error with shallow {ReLU} networks.
\newblock In {\em Proceedings of the {\rm 8}th International Conference on
  Learning Representations ({\rm (}ICLR{\rm )})}, 2020.

\bibitem[LL18]{LiL} Yuanzhi Li and Yingyu Liang.
\newblock Learning overparameterized neural networks via stochastic gradient
  descent on structured data.
\newblock In {\em Proceedings of the {\rm 32}nd Conference on Neural
  Information Processing Systems {\rm (}NeurIPC{\rm )}}, pages 8168--8177,
  2018.

\bibitem[LZ22]{LyZ} Bochen Lyu and Zhanxing Zhu.
\newblock Implicit bias of adversarial training for deep neural networks.
\newblock In {\em Proceedings of the {\rm 10}th International Conference on
  Learning Representations ({\rm (}ICLR{\rm )})}, 2022.

\bibitem[MZ20]{MoZ} Andrea Montanari and Yiqiao Zhong.
\newblock The interpolation phase transition in neural networks: Memorization
  and generalization under lazy training.
\newblock Technical Report 2007.12826 [stat.ML], arxiv e-print, 2020.

\bibitem[NBS18]{NBS} Behnam Neyshabur, Srinadh Bhojanapalli, and Nathan
    Srebro.
\newblock A {PAC}-bayesian approach to spectrally-normalized margin bounds for
  neural networks.
\newblock In {\em Proceedings of the {\rm 6}th International Conference on
  Learning Representations ({\rm (}ICLR{\rm )})}, 2018.

\bibitem[Ngu21]{Ngu} Quynh Nguyen.
\newblock On the proof of global convergence of gradient descent for deep
  {ReLU} networks with linear widths.
\newblock In {\em Proceedings of the {\rm 38}th International Conference on
  Machine Learning {\rm (}ICML{\rm )}}, pages 8056--8062, 2021.

\bibitem[NMM21]{NMM} Quynh Nguyen, Marco Mondelli, and Guido~F.
    Mont{\'{u}}far.
\newblock Tight bounds on the smallest eigenvalue of the neural tangent kernel
  for deep {ReLU} networks.
\newblock In {\em Proceedings of the {\rm 38}th International Conference on
  Machine Learning {\rm (}ICML{\rm )}}, pages 8119--8129, 2021.

\bibitem[OS20]{OyS} Samet Oymak and Mahdi Soltanolkotabi.
\newblock Toward moderate overparameterization: Global convergence guarantees
  for training shallow neural networks.
\newblock {\em IEEE Journal on Selected Areas in Information Theory},
  1(1):84--105, 2020.

\bibitem[SH18]{SoH} Daniel Soudry and Elad Hoffer.
\newblock Exponentially vanishing sub-optimal local minima in multilayer neural
  networks.
\newblock In {\em Proceedings of the {\rm 6}th International Conference on
  Learning Representations ({\rm (}ICLR{\rm )})}, 2018.

\bibitem[SRP{\etalchar{+}}21]{SRP*} Chaehwan Song, Ali Ramezani{-}Kebrya,
    Thomas Pethick, Armin Eftekhari, and
  Volkan Cevher.
\newblock Subquadratic overparameterization for shallow neural networks.
\newblock In {\em Proceedings of the {\rm 35}th Conference on Neural
  Information Processing Systems {\rm (}NeurIPC{\rm )}}, pages 11247--11259,
  2021.

\bibitem[SY19]{SoY} Zhao Song and Xin Yang.
\newblock Quadratic suffices for over-parametrization via matrix {Chernoff}
  bound.
\newblock Technical Report 1906.03593 [cs.LG], arxiv e-print, 2019.

\bibitem[Tro12]{Tro} Joel Tropp.
\newblock User-friendly tail bounds for sums of random matrices.
\newblock {\em Foundations of Computational Mathematics}, 12:389--434, 2012.

\bibitem[Ver12]{Ver} Roman Vershinin.
\newblock {\em Introduction to the non-asymptotic analysis of random matrices},
  chapter~5, pages 210--268.
\newblock Cambridge University Press, 2012.

\bibitem[Ver18]{Ver2} Roman Vershinin.
\newblock {\em High-Dimensional Probability: An Introduction with Applications
  in Data Science}.
\newblock Cambridge University Press, 2018.

\bibitem[VL22]{VlL} Tiffany~J. Vlaar and Benedict~J. Leimkuhler.
\newblock Multirate training of neural networks.
\newblock In {\em Proceedings of the {\rm 39}th International Conference on
  Machine Learning {\rm (}ICML{\rm )}}, pages 22342--22360, 2022.

\bibitem[WDW19]{WDW} Xiaoxia Wu, Simon~S. Du, and Rachel Ward.
\newblock Global convergence of adaptive gradient methods for an
  over-parameterized neural network.
\newblock Technical Report 1902.07111 [cs.LG], arxiv e-print, 2019.

\bibitem[XLS]{XLS} Bo~Xie, Yingyu Liang, and Le~Song.
\newblock Diverse neural network learns true target functions.
\newblock In {\em Proceedings of the {\rm 20}th International Conference on
  Artificial Intelligence and Statistics, {\rm (}AISTATS{\rm )}}.

\bibitem[ZBH{\etalchar{+}}21]{ZBH*} Chiyuan Zhang, Samy Bengio, Moritz Hardt,
    Benjamin Recht, and Oriol Vinyals.
\newblock Understanding deep learning (still) requires rethinking
  generalization.
\newblock {\em Communications of the {ACM}}, 64(3):107--115, 2021.

\bibitem[ZCZG20]{ZCZ*} Difan Zou, Yuan Cao, Dongruo Zhou, and Quanquan Gu.
\newblock Gradient descent optimizes over-parameterized deep {ReLU} networks.
\newblock {\em Machine Learning}, 109(3):467--492, 2020.

\bibitem[ZG19]{ZoG} Difan Zou and Quanquan Gu.
\newblock An improved analysis of training over-parameterized deep neural
  networks.
\newblock In {\em Proceedings of the {\rm 33}rd Conference on Neural
  Information Processing Systems {\rm (}NeurIPC{\rm )}}, pages 2053--2062,
  2019.

\end{thebibliography}

\newcommand{\etalchar}[1]{$^{#1}$}

\appendix

\section{Gradient descent avoids singular points} \label{app:caveat}

Fix $X,W_0,z_0$ such that $\theta_0$ is regular, that is, $W_0X$ does not
contain zero entries. Let $\eta_w,\eta_z\geq 0$ be arbitrary. Our goal in
this section is to show that for Lebesque almost all $y\in\mathbb R^m$, all
points $\theta_\tau\ (\tau =1,2,3,\ldots)$ are also regular.

Due to countable additivity, it is sufficient too prove that
$$
\lambda\of{\set y{\theta_0(y),\theta_1(y),\ldots \theta_{\tau-1}(y)\ \text{are regular},\
\theta_\tau(y)\ \text{is singular}}} =0
$$
for any fixed positive integer $\tau$. By the same token, it is sufficient to show
that for any fixed collection $A_0,A_1,\ldots, A_{\tau-1}$ of 0-1 $(S\times m)$ matrices
and for any fixed $\nu\in [S],\ j\in [m]$,
\begin{equation*}
\begin{split}
&\lambda(\{y | \theta_0(y),\theta_1(y),\ldots, \theta_{\tau-1}(y)\ \text{are regular},\\ &
A_0(y) = A_0,\ A_1(y)=A_1,\ldots,A_{\tau-1}(y)=A_{\tau-1},\ (W_\tau(y)X)_{\nu j}=0\}) =0.
\end{split}
\end{equation*}
But once the activation matrices $A_0,A_1,\ldots,A_{\tau-1}$ are fixed, a
simple induction on $\tau$ shows that all our mappings, including
$(W_\tau(y)X)_{\nu j}$, become polynomial in $y$. Moreover, if we initialize
$y:=f_0$ then $\theta_0$ becomes the global minimum, $\theta_0(y) =
\theta_1(y)=\ldots = \theta_\tau(y)$ and hence $(W_\tau(y)X)_{\nu j}=
(W_0X)_{\nu j}\neq 0$. Hence this polynomial is not identically zero and
therefore the set of its zeros has Lebesque measure 0.

\section{Easy quasirandom properties} \label{app:properties_easy}

In this section we prove that all properties in Section \ref{sec:properties},
except for \eqref{eq:ntk2}, \eqref{eq:ntk1}, hold with probability $\geq
1-(nS)^{-\omega(1)}$; all proofs in this section are routine exercises. As a
preliminary remark, note that the nature of the error probability
$(nS)^{-\omega(1)}$ (along with $m\leq n^{O(1)}$) allows us to freely use the
union bound over sets whose cardinality is polynomial in $n,m,S$.

\medskip \noindent {\bf (\ref{eq:almost_orthogonality})} This is essentially
\cite[Remark 3.2.5]{Ver2} but let us do estimates slightly more carefully.

By the remark just made, we can assume that $j,j'$ are fixed. Let $ x,
 y \sim \mathcal N\of{0, \frac 1nI_n}$ be two independent samples.
Then, due to isotropy, $X^j$ and $X^{j'}$ can be alternately represented as
$X^j\sim \frac{ x}{|| x||},\ X^{j'}\sim \frac{ y}{|| y||}$ and hence
$\absvalue{\langle X^j, X^{j'}\rangle}=\frac{\langle  x,  y \rangle}{||
x||\cdot || y||}$.

Now, for any $i\in [n]$, the random variables $x_i^2-1/n,\ y_i^2-1/n,\
x_iy_i$ are centered and  $||x_i^2-1/n||_{\psi_1},\ ||y_i^2-1/n||_{\psi_1},\
||x_iy_i||_{\psi_1}\leq O(1)$. \cite[Example 2.5.8(i); Lemma 2.7.7; Exercise
2.7.10]{Ver2}. Hence by Bernstein's inequality (\cite[Theorem 2.8.1]{Ver2};
plug $N\mapsto n,\ t\mapsto \frac{\log(nS)}{n^{1/2}}$), with probability
$\geq 1-\exp(-\Omega(\log(nS))^2)\geq 1-(nS)^{-\omega(1)}$ we have $||
x||^2,\ || y||^2\geq 1-\frac{\log(nS)}{n^{1/2}}\geq\frac 12$ and
$\absvalue{\langle  x, y\rangle}\leq \frac{\log(nS)}{n^{1/2}}\leq \wt
O(n^{1/2})$. The bound \eqref{eq:almost_orthogonality} follows.

\medskip\noindent {\bf (\ref{eq:X_norm})} We can treat each value of $k$
individually (and then apply the union bound).

Let us first consider the case $k=m$. By \cite[Theorem 5.39]{Ver} (plug
$N\mapsto m,\ A\mapsto n^{1/2}\cdot X^\top$), there are absolute constants
$c,C>0$ such that for every $t>0$,
$$
\prob{||n^{1/2}\cdot X||\leq C(m^{1/2}+n^{1/2})+t}\geq 1-2\exp(-ct^2).
$$
Set $t\df m^{1/2}\log(nS)$. Then we get that $||X||\leq \wt
O\of{1+\frac{m^{1/2}}{n^{1/2}}}$ with probability $\geq
1-\exp(-\Omega(m\log(nS)^2))\geq 1-(nS)^{-\omega(1)}$.

Let now $k\in [m]$ be arbitrary and $J\in {[m]\choose k}$. Then, plugging in
the above argument $m\mapsto k$, we see that $||X^J||\leq \wt
O\of{1+\frac{k^{1/2}}{n^{1/2}}}$ with probability $\geq
1-\exp(-\Omega(k\log(nS)^2))$, and $\exp(\Omega(k\log(nS)^2))$ dominates the
number ${m\choose k}\leq \exp(k\log m)$ of choices of $J$ (recall again that
$m\leq n^{O(1)}$). Hence we can apply the union bound to complete the proof
of \eqref{eq:X_norm}.

\medskip\noindent {\bf (\ref{eq:X_dual})} This time we have not been able to
find a convenient off-the-shelf inequality so we will do a simple ad hoc
argument using $\epsilon$-nets.

We set $n^\ast \df  n(\log n)^2$ and $\epsilon\df \frac 1m$. By Fact
\ref{fct:net1}, we can find an $\epsilon$-net $\mathcal N$ in $S^{n-1}$ with
$|\mathcal N|\leq\exp(O(n\log n))$. Then it is sufficient to prove that for
any {\em fixed} $\xi\in\mathcal N$ we have
\begin{equation} \label{eq:xi_sigma}
\prob{\forall J\in {[m]\choose n^\ast} ||(X^J)^\top\xi||\geq \frac{2n}{m}}\geq
1-\exp(-\omega(n\log n))
\end{equation}
since then we can apply the union bound over all $\xi\in\mathcal N$, followed
by an application of Fact \ref{fct:net2} (with $\sigma:=n/m$; note that
$||X^J||\leq\wt O(1)$ by the already proven \eqref{eq:X_norm}).

\smallskip
In order to prove \eqref{eq:xi_sigma}, by isotropy we can assume that
$\xi=\of{\begin{array}{c} 1\\ 0\\ \vdots\\ 0 \end{array}}$, that is
$X^\top\xi= X_1^\top$. Let $R\df \frac{n^{1/2}}{m}$, then (by an argument
similar to the one used in the proof of \eqref{eq:almost_orthogonality})
$\prob{|X_{1j}| \leq R}\leq 3Rn^{1/2}\leq\frac{3n}m$ and hence
$\expect{\absvalue{\set j{|X_{1j}|\leq R}}}\leq 3n\leq\frac{n^\ast}{3}$.
Hence, by the plain Chernoff bound,
$$
\prob{\absvalue{\set j{|X_{1j}|\leq R}}\leq \frac{n^\ast}{2}}
\geq 1-\exp(-\Omega(n^\ast)) \geq 1-\exp(-\omega(n\log n)).
$$

On the other hand, this event logically implies that for every $J\in
{[m]\choose n^\ast}$, $(X^J)^\top\xi$ contains at least $\frac{n^\ast}2$
entries $X_{1j}$ with $|X_{1j}|\geq R$. Hence $||(X^J)^\top \xi||\geq
R\of{\frac{n^\ast}{2}}^{1/2}\geq \frac{2n}{m}$. This completes the proof of
\eqref{eq:xi_sigma} and hence of \eqref{eq:X_dual} as well.

\medskip\noindent {\bf (\ref{eq:large_rows})} Fix $\nu\in [S]$. Like in
the proof of \eqref{eq:almost_orthogonality}, for every $i\in [n]$,
$(W_0)_{\nu i}^2-1$ is a centered distribution with $||(W_0)_{\nu
i}^2-1||_{\psi_1}\leq O(1)$. Applying to it Bernstein's inequality
(\cite[Theorem 2.8.1]{Ver2}, plug $N\mapsto m,\ X_i\mapsto (W_0)_{\nu
i}^2-1,\ t\mapsto n/2$), we get
$$
\prob{||(W_0)_\nu||^2 \leq n/2} \geq 1-\exp(-\Omega(n)) \geq 1-(nS)^{-\omega(1)}.
$$

\medskip\noindent {\bf (\ref{eq:z_entries})} The inequality $||W_0||_\infty\leq \wt
O(1)$ is obvious. For $||z_0||_\infty\leq \wt O(1)$, \eqref{eq:psi2} implies
that for sufficiently large $C$,
\begin{equation} \label{eq:bound_on_zeta}
\prob{|\zeta|\leq (\log nS)^C}\geq 1-(nS)^{-\omega(1)}.
\end{equation}

\medskip\noindent {\bf (\ref{eq:z_large})} The bound \eqref{eq:var} on the
variance allows us to conclude that for some $\zeta_0\geq\wt\Omega(1)$,
$\expect{|\zeta|^2\cdot \mathbf 1(|\zeta|\geq\zeta_0)}\geq\wt\Omega(1)$ and
then the bound \eqref{eq:bound_on_zeta} implies $\prob{|\zeta|\geq
\zeta_0}\geq \wt\Omega(1)$. The desired bound \eqref{eq:z_large} now follows
from the plain Chernoff inequality.

\medskip\noindent {\bf (\ref{eq:regular})} is obvious (in fact, it holds with
probability 1).

\medskip\noindent {\bf (\ref{eq:w0x})} Due to isotropy, every individual entry
$(W_0X)_{\nu j}$ is a standard Gaussian.

\medskip\noindent {\bf (\ref{eq:f0})} By the just proven \eqref{eq:w0x}, we also have $||F_0||_\infty \leq
\wt O(1)$ and then, for any fixed $F_0$, the $j$th entry of $f_0=F_0^\top
z_0$ is of the form $a_1\zeta_1+\ldots+a_S\zeta_S$, where $a_\nu\df
(F_0)_{\nu j}$ is fixed, $|a_\nu|\leq\wt O(1)$ and $\zeta_1,\ldots,\zeta_S$
are $S$ independent copies of $\zeta$. Now we only have to apply the
sub-Gaussian Hoeffding inequality \cite[Theorem 2.6.2]{Ver2}.

\medskip\noindent {\bf (\ref{eq:good})} Fix $j\in [m]$. Due to the term +1 in
$RS+1$, we can assume w.l.o.g. that $R\geq \frac 1S$; we can also assume that
$R\leq 1$ as otherwise the bound is trivial. By replacing $R$ with the
nearest rational of the form $2^{-h}\ (h=0,1,2,\ldots)$, we can assume that
$R$ takes on only $\wt O(1)$ different values.  Hence we can apply the union
bound on $R$ and assume that $R\in \left[ \frac 1S,1\right]$ is fixed.

Now, for an fixed $j\in [m]$, the values $(W_0X)_{\nu j}$ are i.i.d. from
$\mathcal N(0,1)$. Hence $\prob{\absvalue{(W_0X)_{\nu j}}\leq R} \leq \wt
O(R)$ and now \eqref{eq:good} follows from the plain Chernoff bound.

\section{NTK matrix at initialization} \label{app:ntk_initialization}

In this section we prove \eqref{eq:ntk2} and \eqref{eq:ntk1}.

Let $w\sim\mathcal N(0,I_n)$. Recall that the {\em limit NTK matrices} (that
we will denote by $H^w(X)$ and $H^z(X)$, respectively) are defined as the
following expectations:
\begin{eqnarray*}
H^w(X) &\df& \indexpect w{(X^\top X)\circ \of{\sigma'(X^\top w) \sigma'(w^\top X)}}\\
H^z(X) &\df& \indexpect w{\sigma(X^\top w) \sigma(w^\top X)}.
\end{eqnarray*}

The first step is to show that \eqref{eq:almost_orthogonality} implies
$\lambda_{\min}(H^w(X)),\ \lambda_{\min}(H^z(X))\geq \Omega(S)$. The argument
is essentially the same as in \cite{NMM} but since we are working in way less
general set-up, we prefer to present a self-contained (and simpler) proof.

\smallskip
The entries $H^w(X)_{jj'},\ H^z(X)_{jj'}$ of those matrices depend only on
$\langle X^j, X^{j'}\rangle$ and this dependence is provided by analytical
(near 0) functions $f^w, f^z$, respectively such that all coefficients in
their Taylor expansions are non-negative.

Indeed, by isotropy we can assume that
$$
X^j= \of{\begin{array}{c} 1\\0 \\ \vdots \\ 0 \end{array}}, \ \ \
X^{j'}= \of{\begin{array}{c} \cos \phi\\ \sin \phi \\ \vdots \\ 0 \end{array}},
$$
where $\phi\df \arccos\of{\langle X^j, X^{j'}\rangle}$. Next, $(w_1,w_2)$ is
distributed as $(r\cos\psi, r\sin\psi)$, where $r$ is the squared
$\chi$-distribution with 2 degrees of freedom, $\psi\sim_R [0,2\pi]$ and
$r,\psi$ are independent. We now compute
$$
(H^w(X))_{jj'} = (X^\top X)_{jj'}\cdot \frac 1{2\pi}\int_{\phi-\pi/2}^{\pi/2} 1d\psi
=\langle X^j, X^{j'}\rangle  \of{\frac 12-\frac{\phi}{2\pi}}
$$
that is,
$$
f^w(\gamma) = \gamma\of{\frac 12-\frac{\arccos \gamma}{2\pi}}
$$
(this computation already appeared several times in the literature). Also,
assuming $\phi\in [0,\pi]$,
$$
(H^z(X))_{jj'} = \expect{r^2}\cdot \frac 1{2\pi}\int_{\phi-\pi/2}^{\pi/2}
\cos(\phi-\psi)d\psi =
\frac{\cos(\phi)\cdot (\pi-\phi) +\sin(\phi)}{2\pi},
$$
that is
$$
f^z(\gamma) = \frac{\gamma(\pi-\arccos(\gamma))+\sqrt{1-\gamma^2}}{2\pi}.
$$
These functions have the following explicit Taylor expansions:
\begin{eqnarray*}
f^w(\gamma)& =& \frac 14\gamma+\frac{1}{2\pi}\cdot\sum_{r=0}^\infty \frac{(2r)!}{4^r\of{r!}^2(2r+1)}\gamma^{2r+2},\\
f^z(\gamma) &=& \frac 1{2\pi} +\frac 14\gamma +\frac 1{48\pi}\gamma^2 + \frac
1{2\pi}\sum_{r=2}^\infty \frac{(2r-3)!!}{(2r)!!} \gamma^{2r}
\end{eqnarray*}
which verifies our claim.

Now, since $H^w(X)$ is obtained from the ordinary kernel matrix $X^\top X$ by
entryism applications of $f^w$, we have
$$
H^w(X) =  \frac 14X^\top X+ \frac{1}{2\pi}\cdot\sum_{r=0}^\infty \frac{(2r)!}{4^r\of{r!}^2(2r+1)}
\underbrace{(X^\top X\circ\cdots\circ X^\top X)}_{2r+2\ \text{times}}
$$
and, by Shur's theorem, all summands in the right-hand side are positive
semi-definite. Hence for every fixed $r\geq 1$, $H^w(X)\succeq
\Omega\of{\underbrace{(X^\top X\circ\cdots\circ X^\top X)}_{2r\
\text{times}}}$. Also, for a fixed $r>0$ all off-diagonal entries in
$\underbrace{(X^\top X\circ\cdots\circ X^\top X)}_{2r\ \text{times}}$ are
bounded as $\widetilde O(n^{r/2})$ (due to \eqref{eq:almost_orthogonality})
which is $o(m)$ if $r>0$ is large enough. On the other hand, all diagonal
entries are still $\Omega(1)$. Hence $\lambda_{\min}\of{\underbrace{(X^\top
X\circ\cdots\circ X^\top X)}_{2r\ \text{times}}}\geq \Omega(1)$ which
completes the proof of $\lambda_{\min}(H^w(X))\geq \Omega(1)$. The argument
for $\lambda_{\min}(H^z(X))\geq\Omega(1)$ is identical.

\medskip
The proof of \eqref{eq:ntk2} is now easy to complete by the matrix Chernoff
inequality. Fix an $X$ satisfying \eqref{eq:almost_orthogonality}. Then $G_0$
is the sum of $S$ independent copies of the random (rank 1) PSD matrix
$\sigma(w^\top X)^\top \sigma(w^\top X)$ while $H^z(X)$ is its expectation.
Moreover,
$$
||\sigma(w^\top X)^\top \sigma(w^\top X)|| = ||\sigma(w^\top X)||^2 \leq
 ||w^\top X||^2.
$$
Let
$$
H^z(X,w) \df \begin{cases} \sigma(w^\top X)^\top \sigma(w^\top X) & \text{if}\
||w^\top X||^2\leq m(\log(nS))^2\\
0 & \text{otherwise}.
  \end{cases}
$$
Then $||H^z(X,w)||\leq \wt O(m)$ and
\begin{equation*}
\begin{split}
||\indexpect w{H^z(X,w)} - H^z(X)|| &\leq \indexpect w{||w^\top X||^2\cdot\mathbf
1\of{||w^\top X||^2\geq m(\log(nS))^2}} \\&\leq  \indexpect w{||w^\top X||^2\cdot\mathbf
1\of{||w^\top X||_\infty \geq \log(nS)}} \leq o(1)
\end{split}
\end{equation*}
(for the last inequality, observe that $w^\top X$ is a tuple of standard
Gaussian, albeit not necessary independent). This in particular implies that
$$
\lambda_{\min}\of{\indexpect w{H^z(X,w)}} \geq\Omega(1),
$$
and now we can apply Matrix Chernoff Inequality (\cite[Theorem 1.1]{Tro};
plug $d\mapsto m,\ X_k\mapsto H^z(X,w),\ R\mapsto \wt O(m),\
\mu_{\min}\mapsto \Omega(S),\ \delta\mapsto 1/2$) to conclude that
\eqref{eq:ntk2} holds with probability $\geq 1-\exp(-\Omega(S/m))\geq
1-\exp(-\wt\omega(1))\geq 1-(nS)^{-\omega(1)}$.

\medskip
The proof of \eqref{eq:ntk1} is tricker, and, as we indicated above, there
probably should be an easier way of doing this.

We first note that $\Gamma_0$ depends only on $z_0$ while $(A_0\ast X)^\top
(A_0\ast X)$ depends only on $X,\ W_0$. Hence we can fix an arbitrary
$\Gamma_0\subseteq [S]$ with $|\Gamma_0|\geq \wt\Omega(S)$ and prove
\eqref{eq:ntk1} conditioned by $\Gamma_0(z_0)=\Gamma_0$. After that we can
simplify our notation by assuming w.l.o.g. that $\Gamma_0=[S]$ (for the
Rademacher initialization this holds automatically anyway).

For any $\Gamma\subseteq [S]$,
$$
||(A_0)_\Gamma\ast X|| \leq ||A_0\ast X|| = ||(A_0^\top A_0)\circ
(X^\top X)||^{1/2} \stackrel{\text{\eqref{eq:shur}}}{\leq} S^{1/2}\cdot ||X||
 \stackrel{\text{\eqref{eq:X_norm}}}{\leq} \wt O\of{S^{1/2}\of{1+
 \frac{m^{1/2}}{n^{1/2}}}}.
$$
Let now
\begin{eqnarray*}
\delta &\df& \lambda_{\min}(H^w(X)) \geq \Omega(1);\\
\sigma &\df& \min\of{\frac{\delta^{1/2}}{3},\ \frac 12} S^{1/2};\\
\epsilon &\df& \frac{\sigma}{2\cdot ||(A_0)_\Gamma\ast X||}\geq
\Omega(m^{-1/2}).
\end{eqnarray*}
Fix an arbitrary $\epsilon$-net $\mathcal N\subseteq S^{m-1}$ of cardinality
$\exp(\wt O(m))$ (by Fact \ref{fct:net1}).
Set also
$$
S^\ast \df \lfloor \frac{n^2S}{(n^2+m)\log(nS)^C} \rfloor,
$$
where $C\geq 1$ is a sufficiently large constant, also to be specified later.
Note that
\begin{equation} \label{eq:S_ast}
S^\ast \geq \wt\omega(m),
\end{equation}
due to $S\geq \wt\omega(m)$ and the assumption $m\leq \wt o(nS^{1/2})$
incorporated in \eqref{eq:ntk1}. Then it is sufficient to show that for any
{\em fixed} $\xi\in\mathcal N$ we have
\begin{equation} \label{eq:fixed_xi}
\prob{\forall\Gamma\in {[S]\choose S-S^\ast}\ ||((A_0)_\Gamma\ast X)\xi||\geq \sigma}
\geq 1-\exp(-\wt\omega(m))
\end{equation}
since after that we can apply the union bound over $\mathcal N$, and then
Fact \ref{fct:net2}.

\smallskip
Before proceeding with the formal proof, let us briefly explain the
predicament we are facing. The bound $\lambda_{\min}(H^w(X))\geq\Omega(1)$ we
have already proven only implies that the {\em expectation} of the random
variable $\xi^\top (X^\top X\circ (\sigma'(X^\top w) \sigma'(w^\top X)))\xi$
is bounded away from zero, and $\xi^\top (A_0\ast X)^\top (A_0\ast X)\xi$ is
a sum of $S$ independent copies of that variable. Since $S\geq \wt\omega(m)$,
we would have been done by a simple application of Chernoff's inequality if
we knew that this random variable does not behave abnormally; say, if we knew
that the {\em probability} that it is separated from 0 is also $\Omega(1)$
(cf. \cite[Assumption 1.2.2]{SoY}).  We can not directly apply Markov's
inequality since we only have an upper bound of $\wt O(m/n)$ on the value of
this random variable. Our way to rule out this pathological situation (i.e.
when the large expectation is made by large values occurring with small
probability) is to apply some ideas from the proof of the Lipschitz
concentration inequality \cite[Theorem 5.1.4]{Ver2}.

\smallskip
Returning to the formal argument, let us first express our random variable in
more compact way:
$$
\xi^\top (X^\top X\circ (\sigma'(X^\top w) \sigma'(w^\top X)))\xi =
||X(\sigma'(X^\top w)\circ \xi)||^2.
$$
Thus,
$$
\indexpect{w}{||X(\sigma'(X^\top w)\circ \xi)||^2}\geq
\lambda_{\min}(H^w(X))=\delta.
$$

For the rest of this section, it will be more convenient to assume that
$w\in_R \sqrt n\cdot S^{n-1}$; since $\sigma'$ is invariant under positive
scalings, this will not change anything. We will denote by $\mu$ the standard
(Haar) measure on $ \sqrt n\cdot S^{n-1}$.

Let
$$
K\df \frac{\delta S}{2S^\ast}
$$
and
$$
\mathcal W \df \set{w\in \sqrt n\cdot S^{n-1}}{||X(\sigma'(X^\top w)\circ \xi)||^2\geq K}.
$$
We split the analysis according to whether $\mathcal W$ is small or large.

\smallskip \noindent
 {\bf Case 1. $\mu(\mathcal W)\leq 1/m$.}

 This case is easy. Since
 $$
 ||X(\sigma'(X^\top w)\circ \xi)||^2\leq ||X||^2\cdot
 ||\sigma'(X^\top w)\circ \xi||^2 \leq ||X||^2 \stackrel{\text{\eqref{eq:X_norm}}}{\leq}
 \wt O\of{1+\frac mn} \leq \wt o(m),
 $$
 we have that $\expect{||X(\sigma'(X^\top w)\circ \xi)||^2\cdot \mathbf{1}(w\in \mathcal W)}\leq
 \wt o(1)$. Hence if we consider the truncated (and scaled by $K$) function
 $$
f(w) \df \min\of{\frac 1K||\sigma'(X^\top w)\circ \xi||^2, 1},
 $$
we will still have $\expect{f(w)}\geq \frac 23\frac{\delta}{K}=\frac
43\frac{S^\ast}{S}$. Noting that $f(w)\in [0,1]$, we have by the plain
Chernoff bound:
$$
\prob{f(w_1)+\ldots+f(w_S)\geq \frac 54S^\ast} \geq 1-\exp(-\Omega(S^\ast)) \stackrel{\text{\eqref{eq:S_ast}}}{\geq}
1-\exp(-\wt\omega(m)).
$$

Now, if we remove from this sum $S^\ast$ terms, it will get decreased by at
most $S^\ast$ and hence
$$
||((A_0)_\Gamma\ast X)\xi||^2\geq K\cdot \sum_{\nu\in\Gamma}f(w_\nu)
\geq \frac 14KS^\ast = \frac 18\delta S,
$$
for any $\Gamma \in {[S]\choose S-S^\ast}$. This completes the analysis of
Case 1.

\smallskip \noindent
 {\bf Case 2. $\mu(\mathcal W)\geq 1/m$.}

 In this case we will be able to achieve our dream goal and show
 that $||X(\sigma'(X^\top w)\circ \xi)||\geq 1$ with probability at least
 1/2.

 For $\rho\geq 0$, let
 $$
\mathcal W_{\rho} \df \set{w\in \sqrt n\cdot S^{n-1}}{\exists w^\ast \in\mathcal
W(||w-w^\ast||\leq\rho)}.
 $$
 Then, by the standard isoperimetric inequality we can fix $\rho\leq \wt O(1)$
 in such a way that $\mu(\mathcal W_\rho)\geq 2/3$. Our goal is to show that
 for all but a negligible fraction of ``bad'' $w\in \mathcal W_\rho$, we have
 $||X(\sigma'(X^\top w)\circ \xi)||\geq 1$.

 In order to identify the first set of ``bad'' points in $\mathcal
 W_\rho$, set first
 $$
\phi\df \frac{\log(nS)^{C_1}}{n^{1/2}},
 $$
 where $C_1\geq 1$ is large enough. Let
 $$
E(w) \df \set{j\in [m]}{||w^\top X^j||\leq\phi},
 $$
 and let $\chi_w\in \{0,1\}^m$ be the characteristic function of this set.
 Then, since $\sum_{j\in [m]}\xi_j^2=1$, we have the estimate
 $$
\indexpect w{||\chi_w\circ\xi||^2} = \sum_{j\in [m]}\xi_j^2\cdot \indprob
w{||w^\top X^j||\leq \phi} \leq \wt O(n^{-1/2}).
 $$
Hence we can choose $\psi\leq\wt O(n^{-1/4})$ such that
$$
\indprob w{||\chi_w\circ \xi||\geq\psi} \leq \frac 1{12},
$$
and this is our first ``bad'' event.

For the second ``bad'' event we need to identify one more quasirandom
property of the data $X$ that in a sense is a uniform version of a dual to
\eqref{eq:good}. Namely, for $R>0,\ w\in \sqrt n\cdot S^{n-1}$ and $X\in
(S^{n-1})^m$, let
\begin{equation} \label{eq:bad_r}
\text{Bad}_R(w,X) \df \absvalue{\set{j\in [m]}{||w^\top X^j||\leq R}}\geq
\log(nS)^2\cdot (mR+1).
\end{equation}
Similarly to the proof of \eqref{eq:good}, for any {\em fixed} $w$ we have
$\indprob X{\text{Bad}_R(w,X)}\leq \exp(-\Omega(\log(nS))^2)$ and, averaging
over $w$,
$$
\prob{\text{Bad}_R(w,X)}\leq \exp(-\Omega((\log(nS))^2)) \leq 1-(nS)^{-\omega(1)}.
$$

On the other hand, if we set
$$
\mathcal B_{R,X} = \set{w\in\sqrt n\cdot S^{n-1}}{\text{Bad}_R(w,X)},
$$
then
$$
\prob{\text{Bad}_R(w,X)} = \indexpect X{\mu(\mathcal B_{R,X})}.
$$
Therefore,
$$
\indprob X{\mu(\mathcal B_{R,X})\leq \wt o(1)} \geq 1-(nS)^{-\omega(1)}.
$$
Let now
$$
\mathcal B_X\df \bigcup_{R>0} \mathcal B_{R,X}.
$$
Then we still have
$$
\indprob X{\mu(\mathcal B_X)\leq \wt o(1)} \geq 1-(nS)^{-\omega(1)}
$$
since (cf. the proof of \eqref{eq:good}) it is sufficient to consider only
$\wt O(1)$ different values of $R$ in this union. Thus we also require that
$$
\mu(\mathcal B_X)\leq \frac 1{12}.
$$
We stress that this is the quasirandom property of the {\em data $X$ only}
and it does {\em not} depend in any way on the actual initialization $W_0$.

The set $\mathcal B_X$ is our second ``bad event'', and now we let
$$
\wt{\mathcal W}_\rho\df \set{w\in \mathcal W_\rho}{||\chi_w\circ \xi||\leq\psi \land
w\not\in \mathcal B_X};
$$
note that $\mu(\wt{\mathcal W}_\rho)\geq \frac 23-2\cdot \frac 1{12}= \frac
12$. We claim that
\begin{equation} \label{eq:W_tilde}
\forall w\in\wt{\mathcal W}_\rho\ ||X(\sigma'(X^\top w)\circ\xi)||\geq 1.
\end{equation}

Indeed, fix $ w\in\wt{\mathcal W}_\rho$ and let $w^\ast\in\mathcal W$ be such
that $||w-w^\ast||\leq\rho\leq \wt O(1)$. Let
$$
d\df \sigma'(X^\top w)- \sigma'(X^\top w^\ast);\ d\in \{-1,0,1\}^m.
$$

We start with the obvious estimate
$$
 ||X(\sigma'(X^\top w)\circ\xi)|| \geq  ||X(\sigma'(X^\top w^\ast)\circ\xi)|| -
 ||X(d\circ\xi)|| \geq K^{1/2}-||X(d\circ\xi)||
$$
(the second inequality holds since $w^\ast\in\mathcal W$). Let now
$D\df\sup(d)$, and split $d$ as $d=d'+d''$, where $d'\df d\circ \chi_w$ is
the part corresponding to $E(w)$ and $d''$ corresponds to $co-E(w)$. Then we
have the bound
\begin{equation} \label{eq:Xdxi_norm}
\begin{split}
||X(d\circ\xi)|| &\leq ||X(d'\circ\xi)||+||X(d''\circ\xi)||
\\&\leq ||X^D||\cdot ||d'\circ\xi|| + ||X^{D\setminus E(w)}||\cdot ||d''\circ \xi||
\\&\leq  ||X^D||\cdot ||\chi_w\circ\xi|| + ||X^{D\setminus E(w)}|| \\&\leq
\psi\cdot ||X^D|| + ||X^{D\setminus E(w)}||.
\end{split}
\end{equation}
Let us estimate $|D|$ and $|D\setminus E(w)|$ (and then we will apply
\eqref{eq:X_norm}).

For any $j\in D$, we have
\begin{equation} \label{eq:opposite_signs}
||(w^\ast -w)^\top X^j||\geq ||w^\top X^j||
\end{equation}
 since $w^\top X^j$ and
$(w^\ast)^\top X^j$ have the opposite sign. Thus, $||(w^\ast -w)^\top
X^D||\geq ||w^\top X^D||$; recalling that $||w^\ast-w||\leq \wt O(1)$, we
obtain
$$
||w^\top X^D|| \leq \wt O\of{||X^D||} \stackrel{\text{\eqref{eq:X_norm}}}{\leq}
\wt O\of{1+\frac{|D|^{1/2}}{n^{1/2}}}.
$$
On the other hand, let
$$
R\df \frac 1m\of{\frac{|D|}{2\log(nS)^2}-1}.
$$
so that the right-hand side in \eqref{eq:bad_r} becomes $\frac{|D|}{2}$.
Since $w\not\in \mathcal B_{R,X}$, there are at least $\frac{|D|}{2}$ indices
$j\in D$ such that $||w^\top X^j||\geq R$ which implies
$$
||w^\top X^D|| \geq \frac 1{\sqrt 2}R|D|^{1/2} \geq \wt\Omega\of{\frac{|D|^{3/2}}{m}} -
\wt O\of{\frac{|D|^{1/2}}{m}}.
$$

Comparing now the upper and lower bounds on $||w^\top X^D||$, we get
$\frac{|D|^{3/2}}{m}\leq\wt O\of{1+\frac{|D|^{1/2}}{n^{1/2}}+
\frac{|D|^{1/2}}{m}}$ which solves to $|D|\leq\wt O\of{m^{2/3}+\frac
m{n^{1/2}}}$. By \eqref{eq:X_norm}, this implies
\begin{equation} \label{eq:d_norm}
||X^D||\leq \wt O\of{1+\frac{m^{1/3}}{n^{1/2}}+\frac{m^{1/2}}{n^{3/4}}} \leq
\wt O\of{1+\frac{m^{1/2}}{n^{3/4}}},
\end{equation}
where the second inequality holds simply because $\frac{m^{1/3}}{n^{1/2}} =
\of{\frac{m^{1/2}}{n^{3/4}}}^{2/3}$.

\smallskip
To bound $|D\setminus E(w)|$, we again use \eqref{eq:opposite_signs} which,
along with the definition of $E(w)$, gives us
$$
||(w-w^\ast)^\top X^{D\setminus E(w)}|| \geq \phi\cdot |D\setminus E(w)|^{1/2}.
$$
On the other hand,
$$
||(w-w^\ast)^\top X^{D\setminus E(w)}|| \leq\rho\cdot ||X^{D\setminus E(w)}||
\stackrel{\text{\eqref{eq:X_norm}}}{\leq} \wt
O\of{1+\frac{|D\setminus E(w)|^{1/2}}{n^{1/2}}}.
$$
If the constant $C_1$ in the definition of $\phi$ is large enough (namely,
exceeds the constant assumed in \eqref{eq:X_norm}), the second term $\wt
O\of{\frac{|D\setminus E(w)|^{1/2}}{n^{1/2}}}$ is dominated by $\phi
|D\setminus E(w)|^{1/2}$ and therefore $\phi |D\setminus E(w)|^{1/2}\leq \wt
O(1)$. Thus $|D\setminus E(w)|\leq \wt O(n)$ and (as always, by
\eqref{eq:X_norm}),
\begin{equation} \label{eq:dew_norm}
||X^{D\setminus E(w)}||\leq \wt O(1).
\end{equation}

Plugging \eqref{eq:d_norm}, \eqref{eq:dew_norm} and $\psi\leq \wt
O(n^{-1/4})$ into \eqref{eq:Xdxi_norm}, we finally conclude that
$$
||X(d\circ\xi)|| \leq \wt O\of{1+\frac{m^{1/2}}{n}}\leq K^{1/2}-1,
$$
provided the constant $C$ in the definition of $S^\ast$ is large enough. This
completes the proof of \eqref{eq:W_tilde}.

\smallskip
The rest is easy. By plain Chernoff bound, $\absvalue{\set{\nu\in
[S]}{w_\nu^\top\in \wt{\mathcal W}_\rho}}\geq S/3$ with probability $\geq
1-\exp(-\Omega(S))\geq 1-\exp(-\wt\omega(m))$. Since $S^\ast\leq o(S)$,
removing any $S^\ast$ entries will still leave us with $\geq\frac S4$ neurons
$\nu\in [S]$ such that $||X(\sigma'(X^\top w_\nu^\top)\circ \xi)||\geq 1$.
This completes the proof of \eqref{eq:fixed_xi} (since $\sigma\leq
\frac{S^{1/2}}{2}$) and \eqref{eq:ntk1}.

\section{NTK matrix at the first layer} \label{app:ntk}

In this section we prove \eqref{eq:ntkh}, assuming that \eqref{eq:bound1},
\eqref{eq:bound2} and \eqref{eq:bound4} hold (for the same $t$).

Let
$$
\Gamma\df \set{\nu\in\Gamma_0}{|(z_t)_\nu|\geq \frac 12\zeta_0}.
$$
Our first task is to check that this $\Gamma$ satisfies the condition
$|\Gamma_0\setminus\Gamma|\leq \wt o\of{\frac{n^2S}{n^2+m}}$ (and hence can
be chosen in \eqref{eq:ntk1}). For that we note that for any $\nu\in
\Gamma_0\setminus \Gamma$, $|(z_t)_\nu-(z_0)_\nu|\geq \frac
12\zeta_0\geq\wt\Omega(1)$ and hence $||z_t-z_0||\geq
\wt\Omega\of{|\Gamma_0\setminus\Gamma|^{1/2}}$. So we only have to check that
\begin{equation} \label{eq:bound_z}
||z_t-z_0|| \leq \min\of{\wt o(S^{1/2}),\ \wt o\of{\frac{nS^{1/2}}{m^{1/2}}}}.
\end{equation}

We can assume (see Remark \ref{rem:dichotomy}) that either
$\frac{\eta_z}{\eta_w}\geq \wt\omega\of{\frac{m^2}{nS}}$ or
$\frac{\eta_z}{\eta_w}\leq \wt O\of{\frac{m^2}{nS}}$.

If $\frac{\eta_z}{\eta_w}\geq \wt\omega\of{\frac{m^2}{nS}}$, we can apply
\eqref{eq:bound4} and use the condition $m\leq\wt o(nS^{1/4})$ in
\eqref{eq:ntkh}.

If on the other hand $\frac{\eta_z}{\eta_w}\leq \wt O\of{\frac{m^2}{nS}}$
then the cases \eqref{eq:wz_case}, \eqref{eq:z_case} are ruled out and hence
\eqref{eq:w_case} must hold. Now we apply \eqref{eq:bound2} to conclude that
$||z_t-z_0||\leq \wt O\of{\frac{m^{3/2}}{n^{1/2}S^{1/2}}}\leq \wt
o\of{\frac{nS^{1/2}}{m^{1/2}}}$, where the last inequality follows from the
calculation $m\leq \wt o\of{(nS^{1/4})^{3/4}\cdot (S)^{1/4}}\leq \wt
o(n^{3/4}S^{7/16})\leq \wt o\of{n^{3/4}S^{1/2}}$. Alternatively, from
\eqref{eq:w_case} we have $\frac{\eta_z}{\eta_w}\leq \wt o\of{\frac Sm}$
which, also by \eqref{eq:bound2}, gives us $||z_t-z_0||\leq \wt o(S^{1/2})$.
Thus in either case we have \eqref{eq:bound_z}.

\medskip
We can now apply \eqref{eq:ntk1} to our particular $\Gamma$, and we get $
\sigma_{\min}((A_0)_\Gamma\ast X) \geq \Omega(S^{1/2})$. Further, since
$(z_t)_\nu\geq \wt\Omega(1)$ whenever $\nu\in\Gamma$, we have
$$
\sigma_{\min}((B_t)_\Gamma\ast X)\geq
\wt\Omega\of{\sigma_{\min}((A_t)_\Gamma\ast X)}.
$$
Thus, all that remains to prove is that $\sigma_{\min}((A_t)_\Gamma\ast
X)\geq \frac 12 \sigma_{\min}((A_0)_\Gamma\ast X)$, and for that it is
sufficient to establish
\begin{equation} \label{eq:a_deviation}
||(A_t-A_0)\ast X|| \leq \wt o(S^{1/2}).
\end{equation}

\medskip
Let
$$
D\df \sup(A_t-A_0) = \set{(\nu,j)\in [S]\times [m]}{\sigma'((W_t)_\nu X^j) \neq
\sigma'((W_0)_\nu X^j)}
$$
and
$$
D^j\df \set{\nu\in [S]}{(\nu,j)\in D}.
$$
Let $J\subseteq [m]$ consist of $\min(m,n)$ data instances $j\in [m]$ with
the maximum value of $|D^j|$ and let $co-J\df [m]\setminus J$. We will bound
$||(A_t-A_0)^J\ast X^J||$ and $||(A_t-A_0)^{co-J}\ast X^{co-J}||$ as $\wt
o(S^{1/2})$ separately.

For the first term, \eqref{eq:X_norm} implies that $||X^J||\leq\wt O(1)$.
Hence, by Shur's inequality \eqref{eq:shur}, it is sufficient to prove that
$\max_{j\in J} |D^j|\leq \wt o(S)$. Fix $j\in J$, and fix $R\geq
\wt\Omega\of{\frac{|D^j|}{S}}$ such that the right-hand side in
\eqref{eq:good} is $\leq\frac{|D^j|}{2}$. Then for at least $\frac{|D^j|}{2}$
values $\nu\in D^j$ we have $|(W_0X)_{\nu j}|\geq R$ and hence
$$
||(W_0)_{D^j} X^j|| \geq \Omega\of{R\cdot |D^j|^{1/2}} \geq \wt\Omega\of{\frac{|D^j|^{3/2}}{S}}.
$$
Also (cf. \eqref{eq:opposite_signs}) $\forall \nu\in D^j (|(W_0)_\nu X^j|\leq
|(W_0-W_t)_\nu X^j|)$. This gives us
\begin{equation} \label{eq:one_column}
||(W_0-W_t)X^j|| \geq  ||(W_0-W_t)_{D^j}X^j|| \geq ||(w_0)_{D^j} X^j|| \geq
\wt\Omega\of{\frac{|D^j|^{3/2}}{S}}.
\end{equation}
On the other hand,
$$
||(W_0-W_t)X^j|| \leq ||W_0-W_t||
\leq ||W_0-W_t||_{\text{\tiny F}} \stackrel{\text{\eqref{eq:bound1}}}{\leq} \wt O(m^{1/2})
\leq \wt o(S^{1/2}).
$$
Comparing these two bounds proves $\max_{j\in J}|D^j|\leq \wt o(S)$ and thus
$||(A_t-A_0)^{J}\ast X^{J}||\leq \wt o(S^{1/2})$.

\smallskip
For the second term $||(A_t-A_0)^{co-J}\ast X^{co-J}||$, we can assume that
$m\geq n$ (and hence $|J|=n$) as otherwise the statement is void. Let $s$ be
such that
$$
\min_{j\in J} |D^j| \geq s \geq \max_{j\in co-J}|D^j|
$$
(it exists due to our choice of $J$). We can now continue
\eqref{eq:one_column} as $||(W_0-W_t)X^j|| \geq
\wt\Omega\of{\frac{s^{3/2}}{S}}\ (j\in J)$ and then conclude
$||(W_0-W_t)X^J||_{\text{\tiny F}}\geq
\wt\Omega\of{\frac{s^{3/2}n^{1/2}}{S}}$.

On the other hand,
$$
||(W_0-W_t)X^J||_{\text{\tiny F}} \stackrel{\text{\eqref{eq:handy}}}{\leq}
||W_0-W_t||_{\text{\tiny F}} \cdot ||X^J|| \stackrel{\text{\eqref{eq:X_norm},
\eqref{eq:bound1}}}{\leq} \wt O(m^{1/2}).
$$
Comparing these two bounds gives us
\begin{equation}\label{eq:bound_on_s}
s\leq \wt O\of{\frac{S^{2/3}m^{1/3}}{n^{1/3}}}.
\end{equation}

Applying now \eqref{eq:shur}, we see that
\begin{equation*}
\begin{split}
||(A_t-A_0)^{co-J}\ast X^{co-J}|| &\leq \of{\max_{j\in co-J}|D^j|}^{1/2} \cdot ||X||
\stackrel{\text{\eqref{eq:X_norm}}}{\leq}
\wt O\of{\frac{s^{1/2}m^{1/2}}{n^{1/2}}} \\&\leq \wt O\of{\frac{S^{1/3}m^{2/3}}{n^{2/3}}}
\stackrel{m\leq \wt o(nS^{1/4})}{\leq} \wt o(S^{1/2}).
\end{split}
\end{equation*}
This completes the proof of \eqref{eq:a_deviation} and hence also of
\eqref{eq:ntkh}.

\section{Gradient does not change radically between steps}
\label{app:between_steps}

In this section we prove \eqref{eq:between_steps} assuming that the bounds
\eqref{eq:bound1}-\eqref{eq:bound4} hold for all $s<t$. By applying another
auxiliary induction\footnote{Recall that due to our convention the gradient
is upper semi-continuous and hence the set of those $t$ for which
\eqref{eq:between_steps} fails is violated is either empty or contains the
minimum element.} we can assume that \eqref{eq:between_steps} also holds for
all $s<t$. In particular, for $s<t$ we may freely use all the conclusions
made in Section \ref{sec:det_part}. Finally, if $t$ is an integer then the
statement is trivially true. Hence we can assume w.l.o.g. that all
inequalities in Section \ref{sec:det_part} hold for $\tau\df\lfloor
t\rfloor$.

\smallskip
Recall that
\begin{eqnarray*}
\nabla\ell(\theta_t) &=& ((B_t\ast X)e_t,\ F_te_t);\\
{\stackrel .\theta}_\tau &=& -(\eta_w,(B_\tau\ast X)e_\tau,\ \eta_zF_\tau e_\tau);\\
\absvalue{\langle \nabla\ell(\theta_\tau), {\stackrel .\theta}_\tau\rangle} &=&
\eta_w||(B_\tau\ast X)e_\tau||^2 + \eta_z||F_\tau e_\tau||^2.
\end{eqnarray*}
Let us introduce the uniform notation
\begin{eqnarray*}
\bar\eta_w &=& \begin{cases} \eta_w& \text{if either \eqref{eq:w_case} or \eqref{eq:wz_case} holds}\\
0 & \text{otherwise}\end{cases}\\
\bar\eta_z &=& \begin{cases} \eta_z& \text{if either \eqref{eq:wz_case} or \eqref{eq:z_case} holds}\\
0 & \text{otherwise}\end{cases}.
\end{eqnarray*}
Then by \eqref{eq:ntkh} and \eqref{eq:ntkg}, the right-hand side of
\eqref{eq:between_steps} gets bounded from below as
\begin{equation} \label{eq:main_bound_pre}
\begin{split}
\absvalue{\langle \nabla\ell(\theta_\tau), {\stackrel .\theta}_\tau\rangle} &\geq
\wt\Omega\of{S^{1/2}||e_\tau||(\bar\eta_w ||(B_\tau\ast X)e_\tau|| + \bar\eta_z
||F_\tau e_\tau||)} \\& \geq \wt \Omega\of{S||e_\tau||^2(\bar\eta_w+\bar\eta_z)}.
\end{split}
\end{equation}

In order to upper bound the left-hand side in \eqref{eq:between_steps}, let
us first collect some useful estimates; in what follows, $s<t$ is arbitrary.

First we have
\begin{equation} \label{eq:W_total}
||W_s-W_0||_{\text{\tiny F}} \leq \wt O(m^{1/2}).
\end{equation}
In the cases \eqref{eq:w_case}, \eqref{eq:wz_case} it follows from
\eqref{eq:bound1}, and in the case \eqref{eq:z_case} -- from
\eqref{eq:bound3} (since $\eta_z\geq \eta_w$ in that case).

The following is the first part\footnote{That part did not use the assumption
$m\leq \wt o(nS^{1/4})$.} of \eqref{eq:bound_z}:
\begin{equation} \label{eq:z0s}
||z_s-z_0|| \leq \wt o(S^{1/2});
\end{equation}
along with \eqref{eq:z_entries} this gives us
$$
||z_s|| \leq \wt O(S^{1/2}).
$$

We can now bound the gradient. Namely, by \eqref{eq:bound_diagonal} we have
\begin{equation} \label{eq:b_s}
||B_s\ast X|| \leq ||z_s||\cdot ||X|| \leq \wt
O\of{S^{1/2}\of{1+\frac{m^{1/2}}{n^{1/2}}}}
\end{equation}
and
\begin{equation*}
\begin{split}
||F_s-F_0||_{\text{\tiny F}} &\leq ||W_sX-W_0X||_{\text{\tiny F}} \leq ||W_s-W_0||_{\text{\tiny F}}\cdot ||X|| \\& \stackrel{\text{
\eqref{eq:W_total}, \eqref{eq:X_norm}}}{\leq} \wt O\of{m^{1/2}+\frac{m}{n^{1/2}}}
\leq \wt O\of{m^{1/2}S^{1/2}}
\end{split}
\end{equation*}
which, along with \eqref{eq:w0x}, implies
$$
||F_s||_{\text{\tiny F}} \leq \wt O\of{m^{1/2}S^{1/2}}.
$$

Now we can also control the evolution from $\tau$ to $t$, both in the
parameter space and in the feature space. Namely, recalling that we have
already proved that $||e_s||$ is decreasing for $s<t$ and that $||e_s||\leq
\wt O\of{m^{1/2}S^{1/2}}$, we have
\begin{equation} \label{eq:W_dif}
\begin{split}
||W_t-W_\tau||_{\text{\tiny F}}& =\eta_w (t-\tau) ||\nabla^w\ell(\theta_\tau)|| \leq \eta_w\cdot
||B_\tau\ast X|| \cdot ||e_\tau|| \\&\leq \wt O\of{\eta_wm^{1/2}S\of{1+
\frac{m^{1/2}}{n^{1/2}}}}
\end{split}
\end{equation}
and thus
\begin{equation} \label{eq:F_diff}
||F_t-F_\tau||_{\text{\tiny F}} \leq ||W_t-W_\tau||_{\text{\tiny F}} \cdot ||X|| \leq \wt O\of{\eta_wm^{1/2}S\of{1+
\frac{m}{n}}}.
\end{equation}
In the $W$-department, we claim that
\begin{equation} \label{eq:W_department}
m\leq \wt o(nS^{1/4}) \Longrightarrow ||(B_t-B_\tau)\ast X|| \leq \wt o(S^{1/2}).
\end{equation}
Since our proof of this fact requires quite new ideas, it is postponed to
Appendix \ref{app:invariant}.

\smallskip
Next, for $s\in [\tau, t)$ we have
$$
\stackrel .e_s = - (\eta_w (B_s\ast X)^\top (B_\tau\ast X) +
\eta_zF_s^\top F_\tau)e_\tau
$$
and hence by the above bounds on $||B_s\ast X||, ||F_s||$,
\begin{equation*}
\begin{split}
||\stackrel .e_s|| &\leq \wt O\of{||e_\tau||\cdot (\eta_w ||B_s\ast X||\cdot
||B_\tau\ast X||+\eta_z ||F_s||\cdot ||F_\tau||)} \\&\leq \wt O\of{S\cdot ||e_\tau||
\cdot\of{\eta_w\of{1+\frac mn}+\eta_zm}}
\stackrel{\text{\eqref{eq:absolute_bounds}}}{\leq} \wt o\of{\frac{||e_\tau||}{S}}.
\end{split}
\end{equation*}
Therefore,
\begin{equation} \label{eq:e_displacement}
||e_t-e_\tau|| \leq \int_\tau^t ||\stackrel .e_s||ds \leq \wt
o\of{\frac{||e_\tau||}{S}}.
\end{equation}

\medskip
Equipped with all this knowledge, we can now proceed to completing
the proof of \eqref{eq:between_steps}.

First, we claim that in \eqref{eq:main_bound_pre} we can now replace
$\bar\eta_w$ with $\eta_w$, i.e. that
\begin{equation} \label{eq:main_bound}
\begin{split}
\absvalue{\langle \nabla\ell(\theta_\tau), {\stackrel .\theta}_\tau\rangle} &\geq
\wt\Omega\of{S^{1/2}||e_\tau||(\eta_w |||(B_\tau\ast X)e_\tau|| + \bar\eta_z
||F_\tau e_\tau||)} \\& \geq \wt \Omega\of{S||e_\tau||^2(\eta_w+\bar\eta_z)}.
\end{split}
\end{equation}
Indeed, we only need to consider the last case \eqref{eq:z_case}. But then
$\bar\eta_z=\eta_z$,
$$
\eta_z||F_\tau
e_\tau||\stackrel{\text{\eqref{eq:ntkg}}}{\geq}
\wt\Omega(\eta_z S^{1/2} ||e_\tau||)
$$
and
$$
\eta_w ||(B_\tau\ast X)e_\tau|| \leq \wt
O\of{\eta_wS^{1/2}\of{1+\frac{m^{1/2}}{n^{1/2}}}||e_\tau||}.
$$
Now, the condition in \eqref{eq:z_case} implies that the former expression
dominates the latter.

Second, let $\delta_t=(\delta_t^w, \delta_t^z)$, where
\begin{eqnarray*}
\delta_t^w &=& (B_t\ast X)e_t - (B_\tau\ast X)e_\tau\\
\delta_t^z &=& F_te_t - F_\tau e_\tau;
\end{eqnarray*}
we bound their contributions separately.

More specifically,
$$
||\delta_t^w|| \leq ||(B_\tau\ast X) (e_t-e_\tau)|| + ||((B_t-B_\tau)\ast X)e_t||
\leq ||B_\tau\ast X|| \cdot ||e_t-e_\tau|| + ||(B_t-B_\tau)\ast X||\cdot||e_\tau||.
$$
We now do some case analysis.

If $m\leq \wt o(nS^{1/4})$, we can apply \eqref{eq:b_s},
\eqref{eq:e_displacement} and \eqref{eq:W_department} to conclude that
$$
||\delta_t^w||\leq \wt o\of{\frac{||e_\tau||}{S^{1/2}}\of{1+\frac{m^{1/2}}{n^{1/2}}} + S^{1/2}||e_\tau||}\leq
\wt o(S^{1/2}||e_\tau||).
$$
Hence
$$
\absvalue{\langle \delta_t^w, {\stackrel .\theta}_\tau \rangle}\leq \eta_w ||\delta_t^w||
\cdot ||(B_\tau\ast X)e_\tau|| \leq \wt o\of{\eta_w S^{1/2} ||e_\tau||\cdot ||(B_\tau\ast X)e_\tau||}
\leq \wt o\of{\absvalue{\langle \nabla\ell(\theta_\tau), {\stackrel .\theta}_\tau \rangle}}
$$
by \eqref{eq:main_bound}.

\smallskip
If, on the other hand, \eqref{eq:z_case} takes place then $\bar\eta_z=\eta_z$
and we bound $||\delta_t^w||$ trivially as
$$
||\delta_t^w|| \leq O\of{(||B_\tau\ast X|| + ||B_t\ast X||)\cdot ||e_\tau||} \leq
\wt O\of{S^{1/2}\of{1+\frac{m^{1/2}}{n^{1/2}}}||e_\tau||}
$$
and then
\begin{equation*}
\begin{split}
\absvalue{\langle \delta_t^w, {\stackrel .\theta}_\tau \rangle}&\leq \wt O\of{\eta_w
S^{1/2}\of{1+\frac{m^{1/2}}{n^{1/2}}}||e_\tau||\cdot ||(B_\tau\ast X)e_\tau||}\\&
\stackrel{\text{\eqref{eq:b_s}}}{\leq} \wt O\of{\eta_w||e_\tau||^2
S\of{1+\frac mn}} \stackrel{\text{\eqref{eq:z_case}}}{\leq} \wt o(\eta_z ||e_\tau||^2S)
\stackrel{\text{\eqref{eq:main_bound}}}{\leq}
\wt o\of{\absvalue{\langle \nabla\ell(\theta_\tau), {\stackrel .\theta}_\tau \rangle}}.
\end{split}
\end{equation*}
This completes the analysis of the contribution of $\delta_t^w$ in
\eqref{eq:between_steps}.

\smallskip
The analysis of $\delta_t^z$ is analogous (and easier):
\begin{equation*}
\begin{split}
||\delta_t^z|| &\leq ||F_\tau(e_t-e_\tau)|| + ||(F_t-F_\tau)e_t||_{\text{\tiny F}} \leq ||F_\tau||
\cdot ||e_t-e_\tau|| + ||F_t-F_\tau||_{\text{\tiny F}}\cdot ||e_\tau|| \\
&\stackrel{\text{\eqref{eq:e_displacement}, \eqref{eq:F_diff}, \eqref{eq:W_dif}}}{\leq} \wt o\of{\frac{m^{1/2}}{S^{1/2}}||e_\tau||+\eta_wm^{1/2}S\of{1+\frac mn}||e_\tau||}
\\& \stackrel{\text{\eqref{eq:absolute_bounds}}}{\leq}
\wt o\of{\frac{m^{1/2}}{S^{1/2}}||e_\tau||}
\end{split}
\end{equation*}
and then
$$
\absvalue{\langle \delta_t^z, {\stackrel .\theta}_\tau \rangle} \leq \wt
o\of{\eta_z\frac{m^{1/2}}{S^{1/2}}||e_\tau||\cdot ||F_\tau e_\tau||}.
$$

If \eqref{eq:wz_case} or \eqref{eq:z_case} takes place then
$\bar\eta_z=\eta_z$ and we are done by \eqref{eq:main_bound}.

If, on the other hand, \eqref{eq:w_case} takes place then we can continue
this estimate as
$$
\absvalue{\langle \delta_t^z, {\stackrel .\theta}_\tau \rangle} \leq \wt o\of{\eta_w
\frac{S^{1/2}}{m^{1/2}}||e_\tau|| \cdot (m^{1/2}S^{1/2})\cdot ||e_\tau||} =
\wt o(\eta_wS||e_\tau||^2),
$$
and we are done again since $\bar\eta_w=\eta_w$.

This completes the proof of \eqref{eq:between_steps}.

\section{Not too many activation changes} \label{app:invariant}

In this section we prove \eqref{eq:W_department}; as noted in the
introduction, our proof uses (a discretized version of) the beautiful
invariant discovered in \cite{ACH,DHL}.

Let us first remind the set-up: we are given a non-integer $t>0$ such that
for all $s<t$ we have all the facts and inequalities proven in Section
\ref{sec:det_part} as well as Appendix \ref{app:ntk}. By continuity, we also
have the seed inequalities \eqref{eq:bound1}-\eqref{eq:bound4} for our chosen
$t$ as well. The bound \eqref{eq:between_steps} is not guaranteed for this
$t$; in fact, this is exactly what we are proving. But it is used only in the
integral form which means that we still have all conclusions from Section
\ref{sec:det_part} and Appendix \ref{app:ntk} for our chosen $t$ as well. We
are specifically interested in \eqref{eq:a_deviation}: $||(A_t-A_0)\ast X||
\leq \wt o(S^{1/2}).$

\medskip
Let us now start the argument. First, we have
$$
||(B_t-B_\tau)\ast X|| \leq ||(\text{diag}(z_t-z_\tau)A_\tau)\ast X|| +
||(\text{diag}(z_t)(A_t-A_\tau))\ast X||.
$$

The estimate of the first term is immediate:
$$
||(\text{diag}(z_t-z_\tau)A_\tau)\ast X|| \stackrel{\text{\eqref{eq:bound_diagonal}}}{\leq}
||z_t-z_\tau|| \cdot ||X|| \stackrel{\text{\eqref{eq:X_norm}}}{\leq} \wt O\of{||z_t-z_\tau||\of{1
+ \frac{m^{1/2}}{n^{1/2}}}}.
$$
The upper bound on $||z_t-z_\tau||$ is obtained via a computation completely
analogous to \eqref{eq:W_dif}:
\begin{equation} \label{eq:z_dif}
||z_t-z_\tau|| = \eta_z(t-\tau)||\nabla^z\ell(\theta_\tau)|| \leq \eta_z||F_\tau||\cdot
||e_\tau|| \leq \wt O(\eta_zmS);
\end{equation}
note for the record (we will need it later) that \eqref{eq:W_dif} and
\eqref{eq:z_dif} also hold for $t\mapsto s,\ \tau\mapsto \lfloor s\rfloor$,
for an arbitrary $s<t$. Hence
$$
||(\text{diag}(z_t-z_\tau)A_\tau)\ast X|| \leq \wt O\of{\eta_zmS\of{1+\frac{m^{1/2}}{n^{1/2}}}}
\stackrel{\text{\eqref{eq:absolute_bounds}}}{\leq} \wt O(1).
$$

\smallskip
It remains to handle the second term, that is
$||(\text{diag}(z_t)(A_t-A_\tau))\ast X||$. We can identify yet another
contribution that we already know how to handle:
\begin{equation*}
\begin{split}
||(\text{diag}(z_0)(A_t-A_\tau))\ast X|| & \stackrel{\text{\eqref{eq:bound_infinity}}}{\leq}
||z_0||_\infty \cdot ||(A_t-A_\tau)\ast X|| \\& \stackrel{\text{\eqref{eq:z_entries}}}{\leq}
\wt O(||(A_t-A_0)\ast X|| + ||(A_\tau-A_0)\ast X||)\\& \stackrel{\text{\eqref{eq:a_deviation}}}{\leq}
\wt o(S^{1/2}).
\end{split}
\end{equation*}
Thus, all that remains to show is
\begin{equation} \label{eq:remains}
||(\text{diag}(z_t-z_0)(A_t-A_\tau))\ast X|| \leq \wt o(S^{1/2}),
\end{equation}
and the difficulty is that we do not have good enough bound on
$||z_t-z_0||_\infty$.

In order to circumvent this difficulty, let
\begin{eqnarray*}
D &\df& \sup(A_t-A_\tau);\\
D_\nu &\df& \set{j\in [m]}{(\nu,j)\in D}
\end{eqnarray*}
and
$$
\Gamma\df \set{\nu\in [S]}{|D_\nu|\leq n^\ast},
$$
where $n^\ast\leq \wt O(n)$ is as in \eqref{eq:X_dual}. We split
$(\text{diag}(z_t-z_0)(A_t-A_\tau))\ast X$ in two parts,
$(\text{diag}(z_t-z_0)(A_t-A_\tau))_\Gamma\ast X$ and
$(\text{diag}(z_t-z_0)(A_t-A_\tau))_{co-\Gamma}\ast X$ and bound their norms
separately.

The first one follows from what we already know:
\begin{equation*}
\begin{split}
||(\text{diag}(z_t-z_0)(A_t-A_\tau))_\Gamma\ast X|| & \stackrel{\text{\eqref{eq:bound_diagonal}}}{\leq}
||z_t-z_0|| \cdot \max_{\nu\in\Gamma} ||X\text{diag}((A_t-A_\tau)_\nu)|| \\&
\stackrel{\text{\eqref{eq:z0s}}}{\leq} \wt o\of{S^{1/2}\cdot\max_{\nu\in\Gamma}||X^{D_\nu}||}
\stackrel{\text{\eqref{eq:X_norm}}}{\leq} \wt o(S^{1/2})
\end{split}
\end{equation*}
(recall that $|D_\nu|\leq \wt O(n)$ for $\nu\in\Gamma$ by our choice of
$\Gamma$).

We bound the last remaining term as
\begin{equation*}
\begin{split}
||(\text{diag}(z_t-z_0)(A_t-A_\tau))_{co-\Gamma}\ast X|| &
\stackrel{\text{\eqref{eq:bound_infinity}}}{\leq} ||(z_t-z_0)_{co-\Gamma}||_\infty \cdot
||(A_t-A_\tau)\ast X|| \\& \stackrel{\text{\eqref{eq:a_deviation}}}{\leq} \wt o\of{S^{1/2}
||(z_t-z_0)_{co-\Gamma}||_\infty}.
\end{split}
\end{equation*}
So it only remains to prove that $||(z_t-z_0)_{co-\Gamma}||_\infty\leq \wt
O(1)$ which, given \eqref{eq:z_entries}, amounts to proving
\begin{equation} \label{eq:z_tau_infinity}
||(z_t)_{co-\Gamma}||_\infty \leq \wt O(1).
\end{equation}
In words: for every particular neuron $\nu\in [S]$ we need to prove the
dichotomy: either its weight on the second layer is small or it changes
activation only on a small number of input data. This is where \cite{ACH,DHL}
steps in.

\medskip
Let us fix a neuron $\nu\in co-\Gamma$. Pick an arbitrary $\wt D_\nu\subseteq
D_\nu$ of cardinality {\em exactly} $n^\ast$. Then we have the following
chain of inequalities:
\begin{equation*}
\begin{split}
||(W_\tau)_\nu|| &\leq \frac mn||(W_\tau)_\nu X^{\wt D_\nu}|| \leq
\frac mn||(W_\tau-W_t)_\nu X^{\wt D_\nu}|| \leq \wt O\of{\frac mn ||(W_\tau-W_t)_\nu||} \\&
 \leq \wt O\of{\frac mn ||W_\tau-W_t||_{\text{\tiny F}}} \leq \wt O\of{\frac mn\eta_w m^{1/2}S
 \of{1+\frac{m^{1/2}}{n^{1/2}}}} \leq \wt O(1).
 \end{split}
\end{equation*}
The first inequality holds since $\sigma_{\min}\of{\of{X^{\wt
D_\nu}}^\top}\geq \frac nm$ by \eqref{eq:X_dual}. The second inequality holds
since for any $j\in\wt D_\nu$, $(W_\tau)_{\nu j}$ and $(W_t)_{\nu j}$ have
opposite signs. The third inequality is true since $||X^{\wt D_\nu}||\leq \wt
O(1)$ by \eqref{eq:X_norm}, the fourth is obvious and the fifth is
\eqref{eq:W_dif}. Finally, the sixth inequality follows from $\eta_w\leq \wt
o\of{\frac n{mS^2}}$ in \eqref{eq:absolute_bounds}. Hence by
\eqref{eq:large_rows} we have
\begin{equation} \label{eq:w_invariant}
||(W_0)_\nu||^2 - ||(W_\tau)_\nu||^2 \geq \wt\Omega(n).
\end{equation}

\medskip
For any $s<t$ such that $\theta_s$ is regular, we have
$$
\frac d{ds}(z_s)_\nu^2 = 2(z_s)_\nu\cdot (\stackrel .z_s)_\nu = -2\eta_z (z_s)_\nu
(F_{\lfloor s \rfloor})_\nu e_{\lfloor s\rfloor}
$$
and
$$
\frac d{ds}||(W_s)_\nu||^2 = 2\langle (W_s)_\nu, (\stackrel .W_s)_\nu\rangle =
-2\eta_w (z_{\lfloor s\rfloor})_\nu (W_s)_\nu X\of{e_{\lfloor s\rfloor}\circ
(A_{\lfloor s\rfloor})_\nu^\top}.
$$
We note that
$$
(W_{\lfloor s\rfloor})_\nu X\of{e_{\lfloor s\rfloor}\circ
(A_{\lfloor s\rfloor})_\nu^\top} = (F_{\lfloor s \rfloor})_\nu e_{\lfloor s\rfloor}.
$$
Hence if we let
$$
R_s\df \eta_w(z_s)_\nu^2 -\eta_z||(W_s)_\nu||^2
$$
then
\begin{equation*}
\begin{split}
||{\stackrel .R}_s|| &\leq \wt O\left(\eta_w\eta_z\left(|(z_{\lfloor s\rfloor})_\nu|
\cdot \absvalue{(W_{\lfloor s\rfloor}-W_s)_\nu X\of{e_{\lfloor s\rfloor}\circ
(A_{\lfloor s\rfloor})_\nu^\top}} \right.\right.
\\& \left.\left. + |(z_{\lfloor s\rfloor} -z_s)_\nu|\cdot
|(F_{\lfloor s \rfloor})_\nu e_{\lfloor s\rfloor}|\right) \right) \\&\leq
\wt O\of{\eta_w\eta_z ||e_{\lfloor s\rfloor}||\cdot\of{||z_{\lfloor s\rfloor}||\cdot
||W_{\lfloor s\rfloor}-W_s||\cdot ||X|| + ||z_{\lfloor s\rfloor}-z_s||\cdot
||F_{\lfloor s\rfloor}|| }}\\& \leq \wt O\of{\eta_w\eta_z ||e_{\lfloor s\rfloor}||\cdot
m^{1/2}S^{3/2}\of{\eta_w\of{1+\frac mn}+\eta_zm}},
\end{split}
\end{equation*}
where for the last inequality we used the bounds from Appendix
\ref{app:between_steps}, as well as \eqref{eq:z_dif}. Let us now integrate
this.

In the case \eqref{eq:w_case} we have $\eta_w\of{1+\frac mn}+\eta_zm \leq \wt
o(\eta_wS)$, and applying \eqref{eq:e_w}, $|R_\tau-R_0|\leq \int_0^\tau
||\stackrel .R_s||ds\leq \wt O(\eta_w\eta_zmS^2)$.

In the two remaining cases \eqref{eq:wz_case}, \eqref{eq:z_case} we have (see
Remark \ref{rem:two_rates})
$\frac{\eta_z}{\eta_w}\geq\wt\omega\of{\frac{m^2}{nS}}$ and then
$\eta_w\of{1+\frac mn}+\eta_zm \leq \wt O\of{\eta_z\of{\frac{nS}{m^2}+ \frac
Sm +m}}$. Hence \eqref{eq:e_z} gives us $|R_\tau-R_0|\leq \wt
O\of{\eta_w\eta_z mS\of{\frac{nS}{m^2}+ \frac Sm +m}}$. Thus in either case
we have the bound
\begin{equation} \label{eq:r_lower}
|R_\tau-R_0| \leq \wt O\of{\eta_w\eta_zS^2\of{m+\frac nm}} \stackrel{\text{\eqref{eq:absolute_bounds}}}{\leq}
\wt o(n\eta_z).
\end{equation}

On the other hand,
\begin{equation} \label{eq:r_upper}
\begin{split}
R_\tau-R_0 &=\eta_z(||(W_0)_\nu||^2 - ||(W_\tau)_\nu||^2) - \eta_w((z_0)_\nu^2 -
(z_\tau)_\nu^2) \\& \stackrel{\text{\eqref{eq:w_invariant}}}{\geq} \wt\Omega(n\eta_z) -\eta_w((z_0)_\nu^2 -
(z_\tau)_\nu^2).
\end{split}
\end{equation}
Comparing \eqref{eq:r_lower} and \eqref{eq:r_upper}, we conclude\footnote{If
$\eta_w=0$ then $D=\emptyset$ and \eqref{eq:W_department} is trivial.} that
$||(z_\tau)_\nu||\leq ||(z_0)_\nu||
\stackrel{\text{\eqref{eq:z_entries}}}{\leq} \wt O(1)$. This concludes the
proof of \eqref{eq:z_tau_infinity}, \eqref{eq:remains} and
\eqref{eq:W_department}.
\end{document}